\documentclass{article}
\usepackage{arxiv}
\usepackage{multirow}
\usepackage[utf8]{inputenc} 
\usepackage[T1]{fontenc}    
\usepackage{newtxtext, newtxmath}
\usepackage{hyperref}       
\usepackage{url}            
\usepackage{booktabs}       
\usepackage{amsfonts}       
\usepackage{nicefrac}       
\usepackage{microtype}      
\usepackage{lipsum}         

\usepackage{algorithm}
\usepackage{algpseudocode}
\usepackage{amsmath}

\usepackage{graphicx}
\usepackage{natbib}
\usepackage{doi}

\usepackage{xr}       
\renewcommand{\undertitle}{}
\date{}
\renewcommand{\headeright}{}

\title{NeuroDyn-EEG: An Interpretable Pre-trained Model for EEG Based on Neural Dynamics}
\author{
  Yi Cui\textsuperscript{1, $\dagger$}, 
  Tong Zhao\textsuperscript{1, $\dagger$}, 
  Jiaxin Lei\textsuperscript{2}, 
  Chuyi Yang\textsuperscript{5}, 
  Yifan Cui\textsuperscript{1}, 
  Ling Zhang\textsuperscript{3, 4, *}, 
  Yuxiang Yan\textsuperscript{1, *}, 
  Bo Hong\textsuperscript{2, *} \\
  \\
  \textsuperscript{1} \textit{Gnosis Neurodynamics Co. Ltd, Room 202, West Area, Building North 1, No. 9 Yard,} \\
  \textit{Shengmingyuan Road, Changping District, Beijing, China} \\
  \textsuperscript{2} \textit{School of Biomedical Engineering, Tsinghua Medicine, Tsinghua University, Beijing, China} \\
  \textsuperscript{3} \textit{Beijing Key Laboratory of Mental Disorders, National Clinical Research Center for Mental Disorders} \\ 
  \textit{and National Center for Mental Disorders, Beijing Anding Hospital, Capital Medical University, Beijing, China} \\
  \textsuperscript{4} \textit{Advanced Innovation Center for Human Brain Protection, Capital Medical University, Beijing, China} \\
  \textsuperscript{5} \textit{University of California, Davis, department of psychology} \\
  \\
  \textsuperscript{$\dagger$} Equal contribution \\
  \textsuperscript{*} Corresponding authors: \texttt{zhangling@ccmu.edu.cn} (L. Zhang), \\
  \texttt{yanyuxiang31@hotmail.com} (Y. Yan), \texttt{hongbo@tsinghua.edu.cn} (B. Hong)
}

\begin{document}

\maketitle

\begin{abstract}
Clinical scalp electroencephalography (EEG) offers a noninvasive window into neural dynamics associated with neuropsychiatric disorders. However, strong discriminative performance from deep representation models does not by itself provide anatomically indexed physiological interpretation. We propose NeuroDyn-EEG, an EEG pretraining framework that incorporates generative priors derived from neural dynamics. The framework combines an extended Jansen-Rit (JR) neural mass model, leadfield-based source-to-scalp projection, and deterministic simulation-based parameter inversion. It is trained on synthetic parameter-EEG pairs sampled within physiologically constrained ranges and estimates 11 families of regional parameters across 90 AAL regions, together with one global parameter, from standard 19-channel EEG. With approximately 2.43 million trainable parameters, NeuroDyn-EEG remains computationally compact.

We evaluate the framework at three levels. First, controlled simulations quantify parameter recovery and degradation under four surrogate-noise conditions. On real resting-state EEG, an inverse-forward closed loop assesses spectral and phase consistency between observed and reconstructed signals. The model preserves aspects of macroscopic rhythmic structure, although recovery varies across parameter families and noise types. Second, across four clinical benchmarks (AD65, PD31, Figshare MDD, and TUAB), NeuroDyn-EEG shows competitive, dataset-dependent performance for case-control or normal-abnormal classification. Among the compared models, it achieves the highest BACC, AUROC, and AUCPR on PD31 and MDD, and the highest BACC with the second-highest AUROC and AUCPR on AD65. Third, post hoc region-wise analyses show that $C_1$, representing local synaptic connectivity from pyramidal cells to excitatory interneurons/stellate cells, has the largest number of regions meeting the prespecified threshold in AD65, whereas the neural-population firing threshold $\theta$ ranks first in Figshare MDD. These distinct parameter-region patterns provide testable hypotheses for subsequent mechanistic studies.

Overall, NeuroDyn-EEG introduces generative constraints derived from neural dynamics and maps scalp EEG directly to anatomically indexed model parameters, providing a computational framework for testing disease-associated dynamical differences. Our code is available at https://github.com/Gnosis-Neurodynamics/NeuroDyn-EEG.

\end{abstract}

\keywords{Neural dynamic equations \and Simulation-based inversion \and Interpretable representation learning \and Clinical EEG \and Generative neural dynamic priors}

\section{Introduction}
Electroencephalography (EEG) records the macroscopic spatiotemporal projections of the electrical activity of neural populations on the scalp surface. Featuring millisecond temporal resolution, low acquisition cost, and bedside noninvasive deployability, EEG has long been applied in clinical scenarios such as epilepsy, cognitive decline, psychiatric disorders, and abnormal EEG screening. Compared with structural MRI or functional neuroimaging with lower temporal resolution, EEG provides a more direct recording of rapidly fluctuating electrophysiological activity \citep{nunez2006electric,buzsaki2012origin}. However, clinical EEG is inherently susceptible to volume conduction, acquisition noise, physiological artifacts, and inter-individual variability, making the stable extraction of discriminative signal features challenging. Against this backdrop, deep learning methods capable of automatically learning task-relevant representations from complex EEG time series have progressively emerged as an important approach for clinical EEG classification and computer-aided analysis \citep{schirrmeister2017deep,lawhern2018eegnet,roy2019deep}.

Early convolutional architectures and compact BCI models demonstrated that end-to-end learning can extract task-specific features from raw EEG \citep{schirrmeister2017deep,lawhern2018eegnet}. Recent EEG foundation models have further leveraged large-scale heterogeneous corpora, masked reconstruction, neural tokenization, or cross-channel Transformer architectures, demonstrating favorable transfer capabilities across various evaluated downstream tasks \citep{jiang2024labram,wang2025cbramod,yang2024brainomni}. Nonetheless, superior classification performance does not inherently equate to physiological interpretability: the latent representations of such models typically lack explicit anatomical indexing and biophysical definitions, making it difficult to delineate which specific brain regions and dynamical parameters exhibit disparities between case and control cohorts. For neurodegenerative and psychiatric disorders, this limitation is particularly prominent, as pathological abnormalities typically involve distributed network dynamics rather than isolated focal alterations \citep{bassett2017network,fornito2015connectomics}.

Neural mass models (NMMs) offer a complementary, mechanism-driven modeling pathway. 
NMMs employ a compact set of macroscopic parameters to describe the mean-field dynamics of excitatory and inhibitory neural populations, thereby establishing a mathematical bridge connecting macroscopic EEG rhythms, network synchrony, and underlying synaptic population processes. 
As a classic paradigm, the Jansen--Rit model describes postsynaptic potential (PSP) interactions between pyramidal cells and excitatory/inhibitory interneuron populations via simplified differential equations \citep{jansen1995electroencephalogram}. 
Subsequent investigations extended it to large-scale networks of coupled cortical columns and noninvasive forward inference, establishing it as a foundational computational tool for investigating cortical circuit dynamics and their macroscopic electrophysiological manifestations \citep{david2003modeling}. 
However, directly inverting whole-brain high-dimensional neural dynamic parameters from sparse scalp EEG channels constitutes a highly ill-posed inverse problem \citep{grech2008review,michel2019eegsource}: 
sparse scalp electrodes linearly mix whole-brain source signals, and drastically distinct underlying parameter combinations can yield virtually identical sensor-level waveforms.

Simulation-based inference (SBI) provides a viable avenue for addressing such problems where explicit likelihood functions are intractable \citep{cranmer2020frontier,sun2024seizure}. 
Its core philosophy involves generating paired ''parameter--observation'' data via a forward simulator, followed by amortized neural networks to infer parameters from observations. 
Mainstream SBI methods frequently approximate posterior distributions via neural density estimation, facilitating uncertainty quantification and calibration analysis \citep{goncalves2020training,tejero2020sbi,boelts2024sbireloaded}. 
In computational neuroscience, related generative modeling paradigms have been widely employed to explore how resting-state brain network patterns emerge from the interplay between local dynamics and structural connectivity \citep{deco2011emerging,sanzleon2013virtualbrain}. 
Nevertheless, in scenarios involving multi-region, high-dimensional whole-brain neural dynamic inversion, posterior distribution estimation typically demands substantial simulation budgets, and the performance of various SBI algorithms fluctuates depending on tasks, evaluation metrics, and computational budgets \citep{lueckmann2021benchmarking}; furthermore, assessing approximate posterior calibration via simulation-based calibration (SBC) requires repeated posterior inference across numerous simulated datasets, imposing heavy computational overhead \citep{talts2018sbc}. 

Inspired by SBI, this study adopts a computationally simpler ``simulation-based parameter inversion'' pathway: training a deterministic inverse mapping network via supervised regression on prior-sampled paired data to yield point estimates rather than full posterior distributions. Under the training distribution jointly defined by the prior and simulator, point estimation optimizing overall risk under MAE loss theoretically converges to the conditional median \citep{bernardo1994bayesian}, which can be viewed as a point summary of the posterior. Grounded on this understanding, we integrate physiologically bounded NMMs, leadfield-based scalp forward projections, and deep representation learning to investigate the feasibility of obtaining anatomically indexed parameter point estimates directly from EEG.

\begin{figure}[t]
\centering
\includegraphics[width=\textwidth]{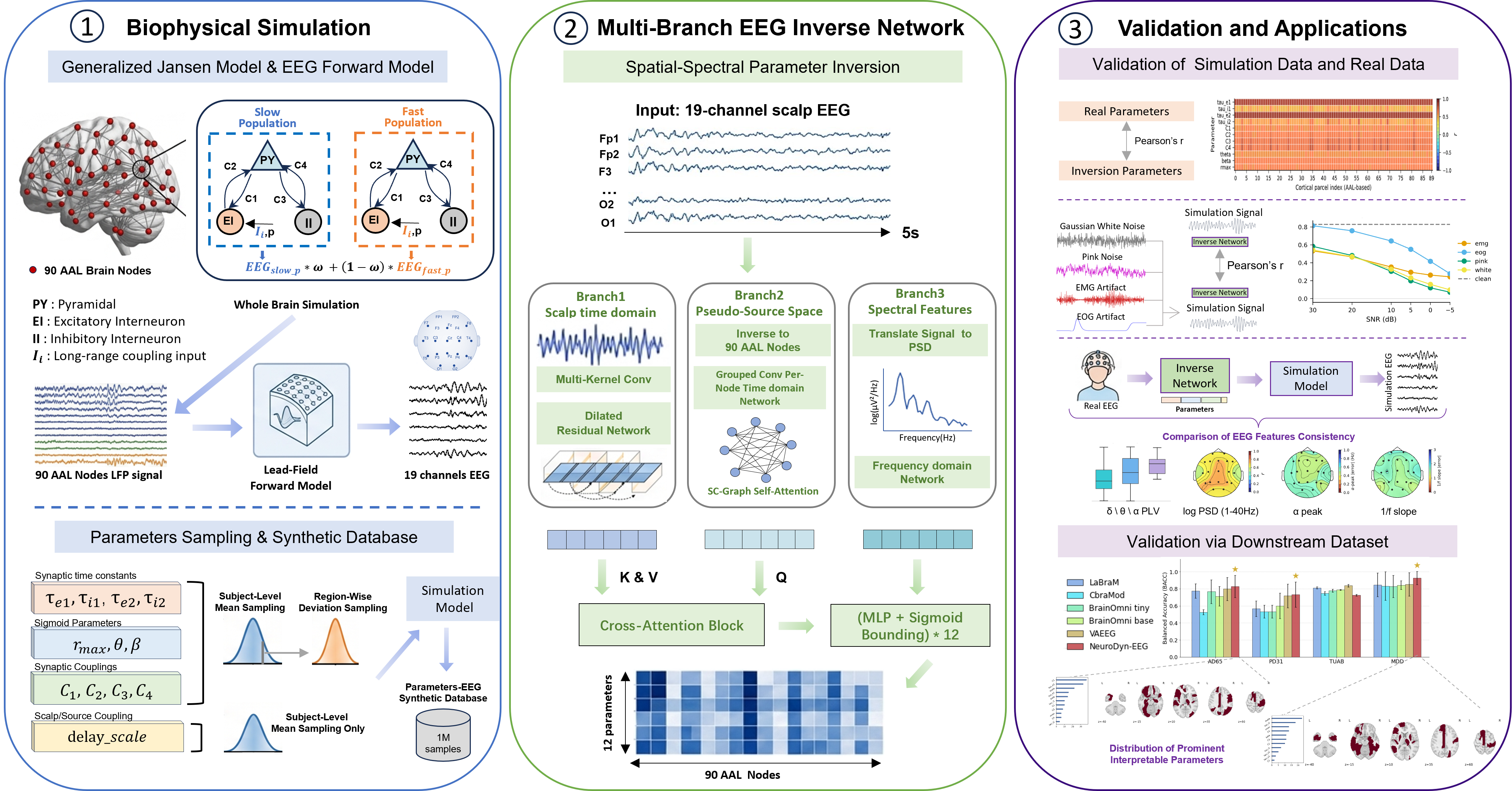}
\caption{\textbf{The three-stage modeling and validation framework of NeuroDyn-EEG.}
(1) \textbf{Whole-brain dynamical simulation}: Constructing a 90-node coupled dual-dynamic neural mass model based on the generalized Jansen model, generating a diverse synthetic ``parameter--scalp EEG'' dataset through physiologically constrained sampling and an intracranial source-to-scalp forward model.
(2) \textbf{Simulation-based multi-branch spatial-spectral parameter inversion model}: Training a multi-branch spatial-spectral inverse network using synthetic scalp EEG-neural dynamic parameter pairs to extract interpretable, region-level parameters from scalp EEG.
(3) \textbf{Systematic evaluation}: Quantifying model performance via parameter recovery, surrogate noise stress-testing, and closed-loop reconstruction; subsequently comparing downstream classification across real clinical cohorts; and finally examining whether inferred parameters exhibit anatomically localizable case--control differences.}
\label{fig:main_framework}
\end{figure}

Guided by this approach, we propose \textbf{NeuroDyn-EEG}---a clinical EEG pretraining framework that deeply incorporates neural dynamic priors. 
As illustrated in Figure~\ref{fig:main_framework}, the framework first drives a 90-node whole-brain coupled dual-dynamic neural mass model, performing large-scale sampling within predefined physiological boundaries to generate intracranial source activities, which are projected to synthetic scalp EEG via forward modeling \citep{grech2008review,michel2019eegsource}. Subsequently, a lightweight multi-branch spatial-spectral inverse network is trained to estimate the JR parameter field distributed across 90 AAL brain regions \citep{tzourio2002automated} from standard 19-channel scalp EEG. Unlike latent representations lacking explicit anatomical indexing, each regional parameter output of NeuroDyn-EEG corresponds to a predefined anatomical brain region and biophysical parameter, enabling rigorous statistical hypothesis testing on specific ``parameter--region'' pairs with multiple comparison corrections. This study addresses three central questions:
(i) What parameter recovery accuracy can the learned inverse model achieve under controlled simulations and diverse surrogate noise conditions? Furthermore, can its inferred biophysical parameters reconstruct observational waveforms that exhibit quantifiable spectral and macroscopic rhythmic consistency with raw signals in an inverse--forward closed loop on real resting-state clinical EEG?
(ii) After incorporating underlying physiological model constraints, how does the model's discriminative performance compare against established deep baseline models across heterogeneous clinical benchmark cohorts?
(iii) Can the inferred system-level parameters yield anatomically organized group-level disparities in case--control cohorts, providing candidate mechanistic hypotheses for subsequent independent validation?

Evaluation results show that NeuroDyn-EEG recovers parameters with parameter-dependent fidelity in controlled simulations and preserves substantial spectral and phase structures in real EEG closed-loop reconstruction; surrogate noise experiments concurrently reveal pronounced noise-type dependencies. Across four clinical benchmarks (AD65, PD31, Figshare MDD, and TUAB), the framework achieves competitive, dataset-dependent downstream performance with approximately 2.43M parameters. Post hoc statistical analyses further reveal that under the current cohort, model, and threshold configurations, \texttt{$C_1$} yields the highest number of threshold-surpassing regions in AD65, whereas \texttt{$\theta$} dominates in Figshare MDD. These spatial distributions qualitatively align with established network neuroscience literature \citep{jeong2004eeg,stam2005eeg,palop2016network,kaiser2015large,mulders2015resting,sanacora1999reduced}, providing testable hypotheses that warrant future independent cohort validation.

\section{Methods}
\subsection{Neural Mass Model}
\label{sec:jr_local_model}
We use the Jansen--Rit (JR) neural mass model \citep{jansen1995electroencephalogram} to describe the average postsynaptic dynamics of a local circuit comprising pyramidal cells, excitatory interneurons, and inhibitory interneurons. The model combines linear synaptic response kernels with a nonlinear transformation from membrane potential to population firing rate. Its state-space representation consists of six first-order ordinary differential equations:
\begin{align}
\dot v_1 &= x_1,
\tag{1.1}\label{eq:jr_v1}\\
\dot x_1 &= \frac{H_e}{\tau_e}S_1(v_2-v_3)
-\frac{2}{\tau_e}x_1-\frac{v_1}{\tau_e^2},
\tag{1.2}\label{eq:jr_x1}\\
\dot v_2 &= x_2,
\tag{1.3}\label{eq:jr_v2}\\
\dot x_2 &= \frac{H_e}{\tau_e}\bigl[p(t)+S_2(v_1)\bigr]
-\frac{2}{\tau_e}x_2-\frac{v_2}{\tau_e^2},
\tag{1.4}\label{eq:jr_x2}\\
\dot v_3 &= x_3,
\tag{1.5}\label{eq:jr_v3}\\
\dot x_3 &= \frac{H_i}{\tau_i}S_3(v_1)
-\frac{2}{\tau_i}x_3-\frac{v_3}{\tau_i^2}.
\tag{1.6}\label{eq:jr_x3}
\end{align}
Here, $v_1$ is the excitatory postsynaptic potential generated by pyramidal-cell firing that drives the local interneuron feedback pathways. The variables $v_2$ and $v_3$ represent the excitatory and inhibitory postsynaptic potential components acting on the pyramidal population. Their difference, $y(t)=v_2(t)-v_3(t)$, defines the regional source signal and determines pyramidal firing. The auxiliary states satisfy $x_k=\dot v_k$ for $k=1,2,3$, and $p(t)$ denotes the external afferent firing-rate input. Time $t$ is continuous in the dynamical equations; simulation outputs are evaluated on a discrete time grid. The parameters $H_e,H_i$ are excitatory and inhibitory synaptic gains, and $\tau_e,\tau_i$ are the corresponding time constants.

Local feedback is mediated by sigmoid functions that include the appropriate connection coefficients:
\begin{equation}
S_k(u)=\frac{c_k^{1}r_{\max}}
{1+\exp\!\left[\beta\left(\theta-c_k^{2}u\right)\right]},
\qquad k\in\{1,2,3\}.
\tag{1.7}\label{eq:jr_sigmoid}
\end{equation}
The parameter $r_{\max}$ is the maximum firing rate of the underlying sigmoid, $\theta$ is its half-activation potential, and $\beta$ controls its slope. The coefficients $c_k^1$ and $c_k^2$ scale the sigmoid output and input, respectively. Specifically, $(c_1^1,c_1^2)=(1,1)$, $(c_2^1,c_2^2)=(\Gamma_2,\Gamma_1)$, and $(c_3^1,c_3^2)=(\Gamma_4,\Gamma_3)$. The effective local connection coefficients are $\Gamma_m=C_{\mathrm{avg}}C_m$, with $C_{\mathrm{avg}}=135$ and $m=1,\ldots,4$; the inferred parameters $C_1$--$C_4$ are dimensionless relative connectivity factors. Thus, both local interneuron pathways are driven by $v_1$, with their respective connection coefficients incorporated in $S_2$ and $S_3$. Potentials and synaptic gains are expressed in mV, firing rates in $\mathrm{s}^{-1}$, and $\beta$ in $\mathrm{mV}^{-1}$. Time constants are converted to seconds for evaluation of the dynamical equations.

\subsection{Coupled Dual-Timescale Neural Mass Model}
\label{sec:neurodyn-dual-kinetic}
Building on the local JR circuit \citep{jansen1995electroencephalogram} and the parallel, dynamically heterogeneous neural-mass framework of David and Friston \citep{david2003modeling}, we construct a coupled network of 90 brain regions. Each region contains slow and fast dynamical branches with distinct synaptic time constants. The dynamics of region $i$ are expressed as six second-order equations:
\begin{equation}
\left\{
\begin{aligned}
\ddot v_1^i &= \frac{H_{e1}}{\tau_{e1}}S_1(y_i)
-\frac{2}{\tau_{e1}}\dot v_1^i-\frac{v_1^i}{\tau_{e1}^{2}},\\
\ddot v_2^i &= \frac{H_{e1}}{\tau_{e1}}\left[I_i(t)+S_2(Z_i)\right]
-\frac{2}{\tau_{e1}}\dot v_2^i-\frac{v_2^i}{\tau_{e1}^{2}},\\
\ddot v_3^i &= \frac{H_{i1}}{\tau_{i1}}S_3(Z_i)
-\frac{2}{\tau_{i1}}\dot v_3^i-\frac{v_3^i}{\tau_{i1}^{2}},\\
\ddot v_4^i &= \frac{H_{e2}}{\tau_{e2}}S_4(y_i)
-\frac{2}{\tau_{e2}}\dot v_4^i-\frac{v_4^i}{\tau_{e2}^{2}},\\
\ddot v_5^i &= \frac{H_{e2}}{\tau_{e2}}\left[I_i(t)+S_5(Z_i)\right]
-\frac{2}{\tau_{e2}}\dot v_5^i-\frac{v_5^i}{\tau_{e2}^{2}},\\
\ddot v_6^i &= \frac{H_{i2}}{\tau_{i2}}S_6(Z_i)
-\frac{2}{\tau_{i2}}\dot v_6^i-\frac{v_6^i}{\tau_{i2}^{2}}.
\end{aligned}
\right.
\tag{2.1}\label{eq:neurodyn-dual-kinetic}
\end{equation}
The states $v_1^i,v_2^i,v_3^i$ belong to the slow branch, and $v_4^i,v_5^i,v_6^i$ are their fast-branch counterparts. The variables $v_1^i,v_4^i$ are the excitatory postsynaptic potentials that mediate local feedback from pyramidal firing. The pairs $v_2^i,v_5^i$ and $v_3^i,v_6^i$ represent excitatory and inhibitory postsynaptic components acting on the pyramidal populations. The parameters $H_{ea},H_{ia}$ and $\tau_{ea},\tau_{ia}$ denote the gains and time constants of branch $a\in\{1,2\}$. Regional indices on local parameters and sigmoid functions are suppressed where the source region is unambiguous.Fast-branch gains are determined by the corresponding time constants \citet{coroneloliveros2026multifrequency}.

The two branches within each region share local connectivity factors and sigmoid parameters, giving $S_1=S_4$, $S_2=S_5$, and $S_3=S_6$, with the input and output scaling defined in Equation~\eqref{eq:jr_sigmoid}. The functions $S_1,S_4$ convert the regional net potential into pyramidal firing rates; $S_2,S_5$ and $S_3,S_6$ describe the excitatory and inhibitory feedback pathways. This shared circuit structure reduces the number of independently varying branch-specific parameters while retaining distinct synaptic response times.

Let $\omega_i$ denote the slow-branch weight. The common potential driving the local interneuron pathways is
\begin{equation}
Z_i(t)=\omega_i v_1^i(t)+(1-\omega_i)v_4^i(t).
\tag{2.2}\label{eq:neurodyn-local-feedback}
\end{equation}
The regional source activity is the weighted sum of the net postsynaptic potentials of the two branches:
\begin{equation}
y_i(t)=\omega_i\left[v_2^i(t)-v_3^i(t)\right]
+(1-\omega_i)\left[v_5^i(t)-v_6^i(t)\right].
\tag{2.3}\label{eq:neurodyn-source-output}
\end{equation}
The inhibitory components enter the source signal with a negative sign. The potential $Z_i$ drives local interneuron feedback, while $y_i$ determines pyramidal firing, regional source output, and long-range input to other regions. the observable rhythms arise from the coupled nonlinear dynamics. This construction provides a multiscale generative model for the $1$--$40$~Hz EEG activity analyzed in this study. 

The background drive is written as $p_i(t)=\mu_i+\widetilde p_i(t)$, where $\mu_i$ is the mean afferent firing rate and $\widetilde p_i(t)$ is its fluctuation. Within each simulated sample, the same sampled sequence was supplied directly to all regions. Combining the background drive with afferent firing-rate fluctuations from other regions gives
\begin{equation}
I_i(t)=\mu_i+\widetilde p_i(t)
+\sum_{j\ne i}K_{ij}^{*}(t)\,
\widetilde S_{1,j}\!\left(y_j(t-d)\right).
\tag{2.4}\label{eq:neurodyn-total-input}
\end{equation}
Here $K_{ij}$ is the structural coupling weight from source region $j$ to target region $i$, and $d$ is the global transmission-delay parameter. The source function $S_{1,j}$ uses the sigmoid parameters of region $j$. Denoting its cumulative historical mean and standard deviation by $\widehat m_j(t)$ and $\widehat\sigma_j(t)$, the centered source firing rate is $\widetilde S_{1,j}(y_j(t-d))=S_{1,j}(y_j(t-d))-\widehat m_j(t)$. For $\widehat\sigma_j(t)>0$ and $2K_{ij}-K_{ij}^2\geq0$, the effective coupling coefficient is
\begin{equation}
K_{ij}^{*}(t)=
\frac{\sigma_{\mathrm{ref}}\sqrt{2K_{ij}-K_{ij}^{2}}}
{\widehat\sigma_j(t)}.
\tag{2.5}\label{eq:neurodyn-effective-coupling}
\end{equation}
This normalization adjusts the scale of source fluctuations, and both dynamical branches receive the same interregional contribution. The variance of the total afferent input depends on the background drive, individual source fluctuations, and their covariances.

To express propagation timing relative to a fixed temporal reference, we reparameterize the delay in the delayed-coupling formulation of David and Friston \citep{david2003modeling} as $d=\tau_{\mathrm{delay,base}}\delta_s$. Here, $\tau_{\mathrm{delay,base}}$ defines the reference
timescale for interregional transmission, and the dimensionless coefficient $\delta_s$ controls the delay magnitude relative to this reference. Accordingly, the source activity entering the interregional coupling term is evaluated at $t-\tau_{\mathrm{delay,base}}\delta_s$.
This formulation provides a normalized coordinate for varying the temporal offset of interregional inputs, while the local synaptic time constants govern the postsynaptic response kinetics within each region. Parameter definitions, fixed simulation settings, and the hierarchical sampling procedure are provided in Supplementary Section~\ref{sec:supp_sampling}.

\subsection{Intracranial Forward Projection Model}
\label{sec:forward_model}
This study utilizes the MNE standard head model \citep{gramfort2014mne} to construct the forward propagation model from intracranial source space to scalp electrode space, following standard EEG source imaging procedures: defining head models and leadfield matrices prior to source-to-sensor projection \citep{grech2008review,michel2019eegsource}. First, a 3D voxel grid is extracted from the MNE template to obtain spatial coordinates of all candidate source points. Subsequently, these voxel coordinates are co-registered with the centroid coordinates of 90 standard brain regions defined by the Automated Anatomical Labeling (AAL) atlas \citep{tzourio2002automated}. 

To identify representative source dipoles for each region, Euclidean distances between each AAL centroid and all intracranial voxels are computed. For each AAL region, a set of $K=10$ nearest neighbor voxels is selected as the representative node set for that anatomical structure. Corresponding leadfield vectors are extracted from the complete leadfield matrix, which characterizes the scalp potential distribution generated by a unit current dipole at each intracranial source location. Finally, index mapping retrieves forward propagation coefficients associated with the 90 AAL regions, forming a forward model matrix. This matrix maps the anatomically defined source space to the 19 standard scalp electrode locations of the International 10--20 system \citep{jasper1958ten}. The conversion from intracranial source activity to scalp EEG is realized via the dot product between AAL regional signals and the forward projection matrix.

\subsection{Simulation-Based Multi-Branch Spatial-Spectral Inversion Architecture}
\label{sec:sbi_multihead}

\begin{figure}[t]
    \centering
    \includegraphics[width=0.95\textwidth]{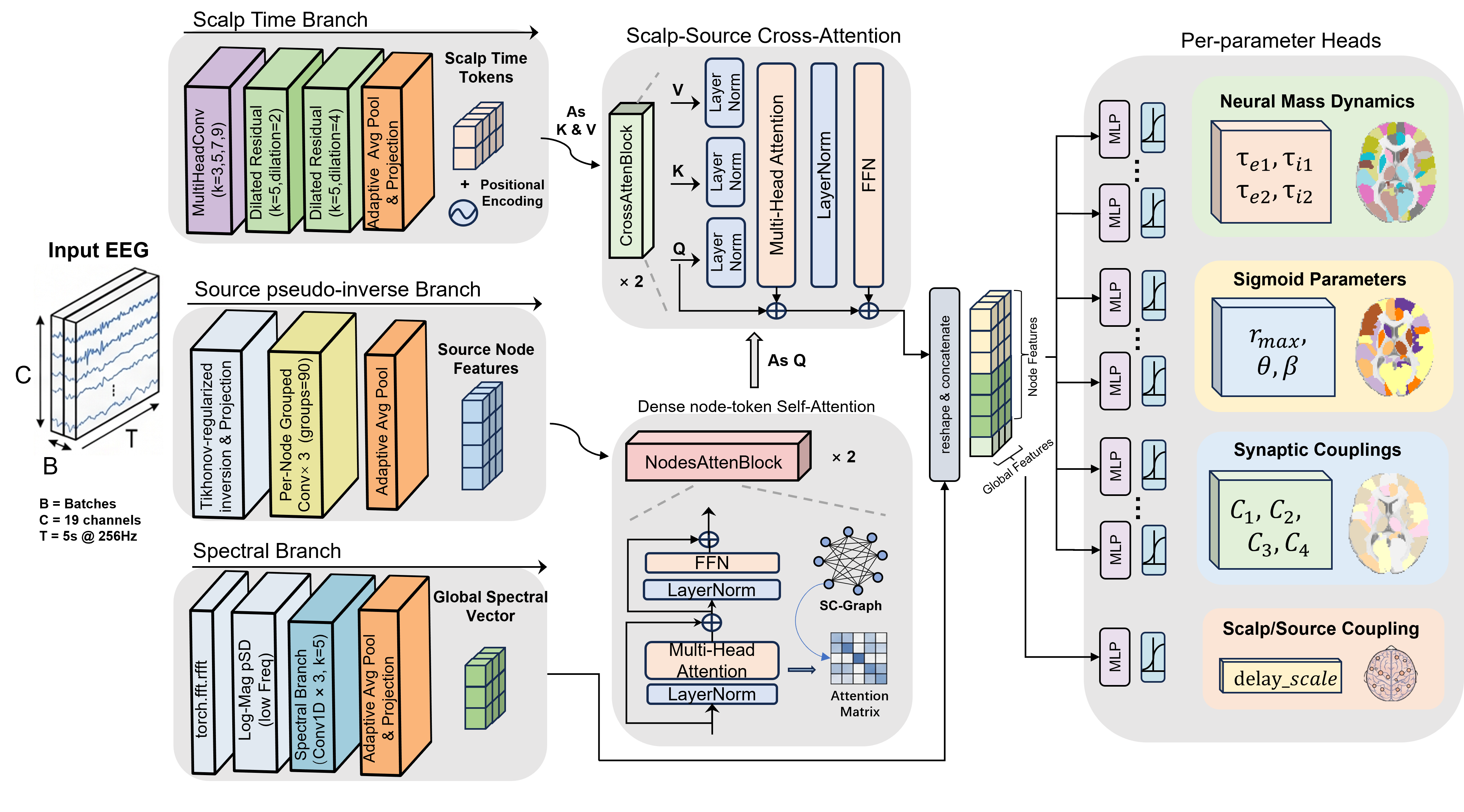}
    \caption{\textbf{NeuroDyn-EEG parameter inversion network architecture.} The framework adopts a multi-branch spatial-spectral design: The scalp temporal branch extracts long-range temporal features via multi-scale convolutions and dilated residual blocks; the source-space branch projects scalp EEG onto 90 source nodes via the leadfield pseudoinverse; and the spectral branch encodes log-magnitude spectral features. Node self-attention and scalp--source cross-attention integrate these representations, and parameter-specific heads produce regional estimates across 90 AAL brain regions together with one global delay coordinate under the specified output constraints.}
    \label{fig:model_framework}
\end{figure}

We train a deterministic inverse network $g:\mathbf X\mapsto\widehat{\boldsymbol\phi}$ using paired parameter--EEG samples generated by the coupled JR model and scalp forward projection. Parameters are sampled from the simulation prior, and supervised regression learns a direct mapping from the resulting EEG signals to parameter point estimates.

The inverse problem involves 11 regional parameter families across 90 brain regions and one global delay coordinate inferred from 19 scalp channels. Under the joint distribution defined by the prior and simulator, componentwise conditional medians minimize the expected absolute error for an unrestricted predictor \citep{bernardo1994bayesian}. The implemented network provides a constrained approximation through its shared feature representation and parameter-specific output mappings. Its multi-branch architecture integrates temporal features, leadfield-based source representations, and attention across node tokens (Figure~\ref{fig:model_framework}). The network contains approximately 2.43M trainable parameters; layer configurations and tensor dimensions are provided in Supplementary Table~\ref{tab:supp_neurodyn_architecture}.

\textbf{Scalp Temporal Branch.} This branch comprises four parallel 1D convolutional layers with kernel sizes of 3, 5, 7, and 9 to capture multi-scale temporal dynamics. Each branch is processed with Batch Normalization \citep{ioffe2015batch} and LeakyReLU activation \citep{maas2013rectifier} before concatenation along the channel dimension. Two cascaded {dilated residual blocks expand the receptive field without downsampling, providing adequate temporal context for long-range dependent parameters (e.g., $\mathrm{delay\_scale}$ and time constants). The resulting feature sequence is pooled and projected to dimension $d_{\text{model}}=192$, serving as key-value memory for downstream cross-attention.

\textbf{Source-Space Pseudoinverse Branch.} Accounting for scalp volume conduction mixing and the ill-posed nature of source localization \citep{grech2008review,michel2019eegsource}, we employ the Tikhonov-regularized pseudoinverse $L^{+}\in\mathbb{R}^{90\times19}$ (ridge parameter $10^{-3}$) of the leadfield matrix $L\in\mathbb{R}^{19\times90}$ to project scalp signals into a 90-dimensional source space. Inverted source signals are independently encoded via grouped convolutions to construct localized representations indexed by AAL nodes.

\textbf{Spectral Branch.} Given that NMMs generate characteristic oscillations across $\delta/\theta/\alpha/\beta$ bands, we incorporate an auxiliary spectral branch: computing the fast Fourier transform and log-magnitude spectrum for each channel, encoded into a 64-dimensional global spectral feature via a lightweight convolutional network. This feature provides complementary statistical cues for spectral-shape-dependent parameters.

\textbf{Node Self-Attention and Scalp--Source Cross-Attention.} Pseudoinverse features of each source node are concatenated with a 32-dimensional learnable node embedding and projected into $d_{\text{model}}=192$ node tokens. Along with a global query token, these are fed into a 2-layer self-attention module to learn input-dependent inter-nodal interactions. Node tokens then serve as queries to attend to the temporal memory from the scalp branch via a 2-layer cross-attention module, capturing data-dependent associations between electrode temporal features and node representations.

\textbf{Parameter-Specific Prediction Heads and Range Constraints.}
The fused node and spectral features are mapped to regional parameters by independent prediction heads, with each head sharing its weights across the 90 AAL regions.
A separate global head predicts the delay coordinate from the global token and spectral features.
Parameter-specific sigmoid mappings constrain the outputs to configured ranges, with optional margins allowing limited extension beyond the reference bounds.
For both slow and fast branches, each inhibitory time constant is additionally constrained between a predefined lower bound and 2.4 times its corresponding predicted excitatory time constant.
These differentiable output mappings incorporate parameter-wise range constraints and dependencies between excitatory and inhibitory time constants directly into the prediction heads. The complete parameterization is described in Supplementary Section~\ref{sec:supp_training}.

\subsection{Parameter Sampling and Synthetic Dataset Generation}
\label{sec:param_sampling}
The coupled dual-timescale JR model generates the regional source activity. Interregional coupling is specified by a fixed structural connectivity matrix obtained by averaging DTI-derived tractography matrices from 88 healthy participants \citep{skoch2022human}. Each simulated sample is defined by 11 parameter families varying across the 90 AAL regions and one global delay coordinate $\delta_s$ (\texttt{delay\_scale}). Sampling bounds and fixed simulator settings are listed in Supplementary Tables~\ref{tab:supp_sampling} and \ref{tab:supp_fixed}.

Directly sampled regional parameters follow a two-level truncated normal scheme. A sample-level latent location is drawn within the specified interval, followed by conditionally independent regional draws using the same latent location and a smaller parent-Gaussian dispersion. The shared latent variable induces interregional dependence within each parameter family. The distribution parameters and their relation to the truncated moments are specified in Supplementary Section~\ref{sec:supp_sampling} and Table~\ref{tab:supp_sampler_hp}. Inhibitory time constants are sampled conditionally on their excitatory counterparts, with support
\[
\max(\tau_{ea},\ell_{ia})\leq\tau_{ia}
\leq\min(h_{ia},\kappa\tau_{ea}),
\qquad a\in\{1,2\},\quad\kappa=\frac{12}{5}.
\]
The global coordinate is drawn once per sample from $\delta_s\sim\mathrm{Uniform}(0,1)$ and converted to the nominal physical delay $d=(10~\mathrm{ms})\delta_s$.

Regional source signals are projected to the 19 standard 10--20 scalp channels using the leadfield matrix (Section~\ref{sec:forward_model}). The resulting scalp EEG samples are represented as 5-second segments at 256~Hz. The synthetic corpus contains $S=10^6$ paired parameter--EEG samples, with $\mathbf X\in\mathbb R^{19\times1280}$, regional targets $\boldsymbol\phi_{\mathrm{regional}}\in\mathbb R^{11\times90}$, and one scalar target $\delta_s$ per sample.

\subsection{Training Implementation and Objectives}
Preprocessing comprises temporal demeaning within each channel, spatial demeaning across channels at each time point, and amplitude screening, as detailed in Supplementary Section~\ref{sec:supp_training}. The regional target families are
\[
\mathcal P_{\mathrm{regional}}=
\{\tau_{e1},\tau_{i1},\tau_{e2},\tau_{i2},\theta,\beta,r_{\max},C_1,C_2,C_3,C_4\}.
\]
Training minimizes a weighted mean absolute error over these 11 families and the global delay coordinate. For a single sample, the objective is
\begin{equation}
\mathcal L=
\sum_{k\in\mathcal P_{\mathrm{regional}}}w_k
\left[\frac{1}{N}\sum_{i=1}^{N}
\left|\widehat\phi_k^{,i}-\phi_k^{,i}\right|\right]
+w_{\mathrm{delay}}\left|\widehat\delta_s-\delta_s\right|,
\qquad N=90,\quad w_{\mathrm{delay}}=1.
\tag{5}\label{eq:loss}
\end{equation}
Here $\widehat\phi_k^{\,i}$ and $\phi_k^{\,i}$ are the predicted and sampled target values for parameter family $k$ in region $i$, and $\widehat\delta_s,\delta_s$ are the predicted and sampled global delay coordinates. Losses are evaluated in the stored target coordinates and averaged over the mini-batch. Regional loss weights $w_k$ are listed in Supplementary Table~\ref{tab:supp_weights}. The model is trained by supervised regression to produce deterministic parameter point estimates.

\subsection{Simulation and Real-Data Validation Protocols}
\label{sec:sim_validation_methods}
We evaluate inversion quality across simulated corpora and real resting-state EEG cohorts, alongside surrogate artifact stress-testing.

\textbf{(1) Parameter Recovery (Simulation):} On an independent synthetic test set, model predictions $\hat{\boldsymbol{\phi}}=g(\mathbf{X})$ are evaluated against ground-truth $\boldsymbol{\phi}$. Linear recovery quality is quantified via Pearson correlation coefficients ($r$) at both sample-level marginals (Figure~\ref{fig:sim_validation_all}\textbf{a}) and regional AAL distributions (Figure~\ref{fig:sim_validation_all}\textbf{b}).

\textbf{(2) Noise Robustness (Simulation):} Four surrogate artifact classes---white Gaussian noise, $1/f$ pink noise, electromyographic (EMG)-like, and electrooculographic (EOG)-like noise---are injected into clean synthetic segments across $\text{SNR} \in \{30, 20, 10, 5, 0, -5\}$\,dB (Supplementary Section~\ref{sec:supp_noise}). Pearson $r$ between predictions and ground truths is tracked across noise levels to evaluate parameter-specific resilience.

\textbf{(3) Real EEG Closed-Loop Consistency:} Resting-state 19-channel EEG from 369 healthy individuals (collected at Beijing Anding Hospital, Capital Medical University, Approval Number:2021-86,2024-205) is inverted into parameters $\hat{\boldsymbol{\phi}}=g(\mathbf{X}_{\mathrm{obs}})$ and fed back into the forward NMM simulator to generate reconstructed waveforms $\tilde{\mathbf{X}}$. Spectral consistency (1--40 Hz) is assessed via: (i) Pearson correlation of log Welch PSD ($r_{\log}$); (ii) absolute $\alpha$-peak frequency error (7--13 Hz); and (iii) absolute error in aperiodic $1/f$ spectral slope. Temporal consistency is evaluated using Phase-Locking Value (PLV) across $\delta$ (1--4 Hz), $\theta$ (4--8 Hz), and $\alpha$ (8--13 Hz) bands (Figure~\ref{fig:sim_validation_all}\textbf{d}--\textbf{e}). The $\beta$ band is excluded from PLV analysis due to susceptibility to aperiodic noise and muscle artifacts.

\section{Downstream Experiments}

\subsection{Baseline Models}
We benchmark \textbf{NeuroDyn-EEG} against four representative deep EEG representation models:
\begin{itemize}
    \item \textbf{LaBraM:} A large-scale neural Transformer trained via vector-quantized neural spectrum prediction and masked reconstruction on heterogeneous EEG corpora \citep{jiang2024labram}.
    \item \textbf{CBraMod:} An EEG foundation model utilizing a Criss-Cross Transformer and asymmetric conditional positional encodings for channel adaptation \citep{wang2025cbramod}.
    \item \textbf{BrainOmni:} A unified foundation model for EEG/MEG representation learning combining neural tokenization and sensor-aware encoding \citep{yang2024brainomni}.
    \item \textbf{VAEEG:} A VAE-based EEG representation model incorporating residual blocks and frequency-band decomposition \citep{zhang2023vaeeg}.
\end{itemize}

\subsection{Downstream Datasets}
Evaluation spans four public clinical EEG datasets covering diverse neuropsychiatric conditions:
\begin{itemize}
    \item \textbf{AD65 (Alzheimer's Disease):} Resting-state EEG from 65 subjects (36 AD, 29 healthy controls), recorded across 19 channels at 500 Hz for binary AD classification.
    \item \textbf{PD31 (Parkinson's Disease):} Resting-state EEG from 31 subjects (15 PD, 16 healthy controls), recorded across 32 channels at 512 Hz.
    \item \textbf{Figshare MDD (Major Depressive Disorder):} Resting-state EEG from 65 subjects (35 MDD, 30 healthy controls) recorded across 20 channels at 256 Hz, containing eyes-open, eyes-closed.
    \item \textbf{TUAB (Temple University Abnormal EEG Corpus):} A large-scale clinical corpus labeled by clinical experts as normal or abnormal, serving as a general benchmark for clinical abnormality detection.
\end{itemize}

\subsection{Preprocessing and Harmonization Strategies}
\label{sec:downstream_preproc}
To accommodate varying channel layouts, sampling rates, and normalization protocols across baseline models, preprocessing pipelines were configured strictly according to their official implementations and publications (Supplementary Section~\ref{sec:supp_preprocess}). For NeuroDyn-EEG, clinical EEGs were harmonized into standard 19-channel formats matching the pretraining setup. All models share identical subject-level data splits and evaluation metrics to ensure fair comparison.

\subsection{Neural Dynamic Parameter Inference and Univariate Case--Control Analysis}
\label{sec:jr_param_stats_methods}
To investigate whether inferred parameters exhibit anatomically organized case--control differences, we conduct group-level statistical testing on inferred JR parameters rather than relying on unsupervised latent clustering. Diagnostic labels and cohorts remain identical to downstream classification setups. For each subject, trial-averaged parameter estimates $\hat{\boldsymbol{\theta}}$ are obtained for each AAL region (excluding the global scalar \texttt{delay\_scale}). Independent two-tailed Welch's $t$-tests \citep{welch1947generalization} evaluate differences between case and control groups across all region--parameter pairs. 

For each dataset, the 990 regional tests (11 parameter families $\times$ 90 AAL regions) plus 1 global test for \texttt{delay\_scale} constitute a unified family of hypotheses, corrected for False Discovery Rate (FDR) using the Benjamini--Hochberg (BH) procedure ($q_{\mathrm{BH}}$) \citep{benjamini1995controlling}. A significance threshold of $q_{\mathrm{BH}} < 0.001$ is adopted to identify candidate regions with pronounced group differences. Macro-level effects are summarized by ranking parameter families by the count of significant regions and rendering brain surface topographies on the standard AAL-90 template. Complete statistics ($t$-values, raw $p$, $q_{\mathrm{BH}}$, effect sizes, sample sizes) are detailed in Supplementary Section~\ref{sec:supp_jr_stats}.

\section{Results}
\label{sec:results}

\subsection{Simulation-Based Inversion Benchmarks}
\label{sec:sim_validation_results}

\begin{figure}[t]
    \centering
    \includegraphics[width=\textwidth]{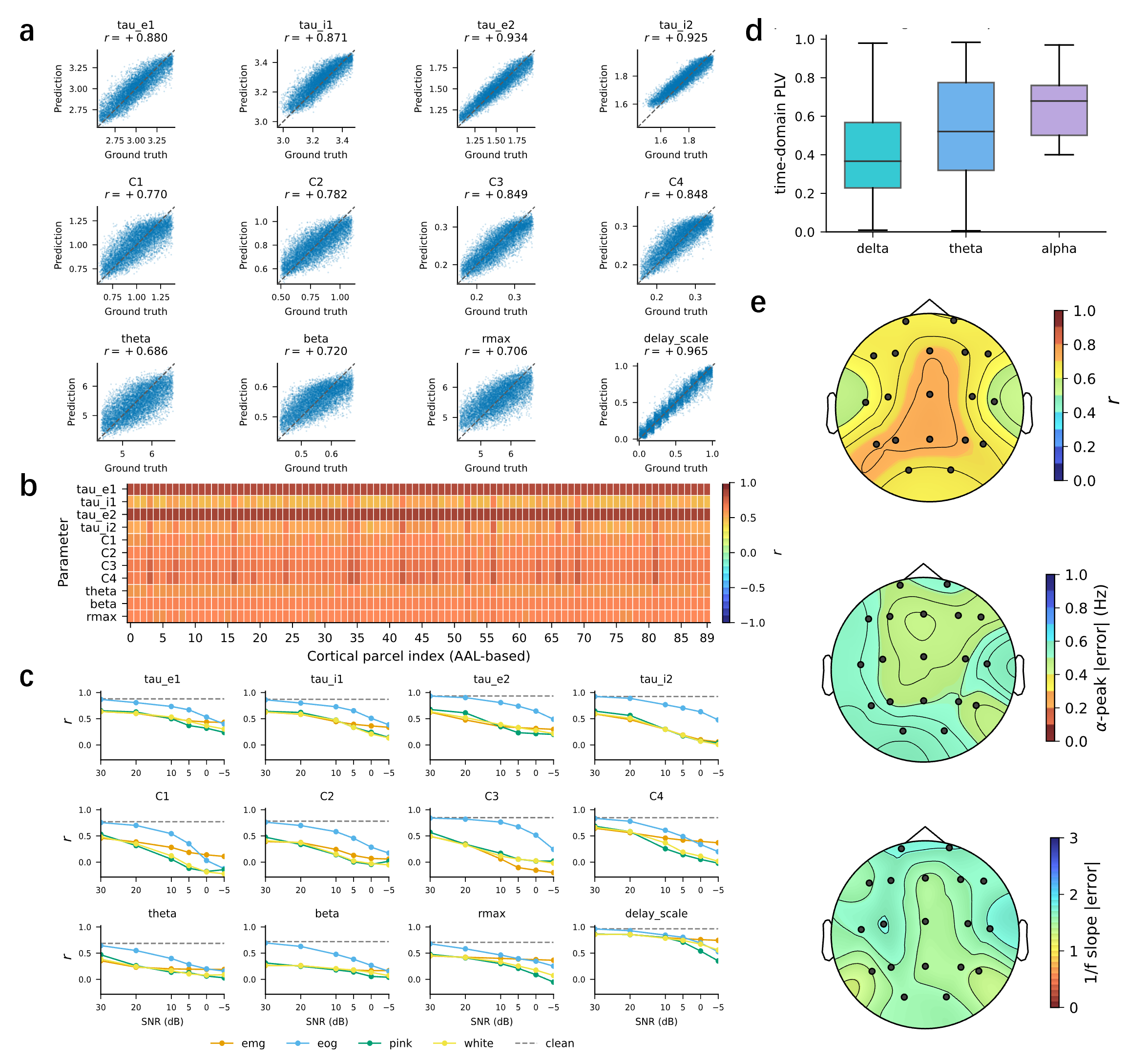}
    \caption{\textbf{Simulation parameter recovery and real EEG closed-loop evaluation.}
    (\textbf{a}) Sample-level predicted vs. true values across 12 targets on independent synthetic test data; dashed line indicates ideal identity ($y=x$), with Pearson $r$ annotated.
    (\textbf{b}) Heatmap of region-wise Pearson $r$ across AAL cortical parcels.
    (\textbf{c}) Parameter-wise Pearson $r$ curves under four surrogate noise types (white noise, pink noise, EMG, EOG) across SNR 30 to $-5$\,dB, referencing clean baseline performance.
    (\textbf{d}) Boxplots of multi-band Phase-Locking Value (PLV) in real resting-state EEG closed-loop reconstruction ($\delta$, $\theta$, $\alpha$ bands).
    (\textbf{e}) Topographic maps of channel-wise reconstruction fidelity on real EEG ($n=369$); Top: 1--40 Hz log PSD Pearson $r$; Middle: absolute $\alpha$-peak frequency error (Hz); Bottom: $1/f$ aperiodic slope error. Black dots indicate standard 10--20 electrode positions.}
    \label{fig:sim_validation_all}
\end{figure}

\textbf{Parameter Inversion under Ideal Observations:} As shown in Figure~\ref{fig:sim_validation_all}(\textbf{a}), the 12 target parameters achieve an average Pearson correlation of $r = 0.83$ on the independent simulation test set. The global conduction delay scaling factor $\mathrm{delay\_scale}$ achieves the highest recovery ($r = 0.97$), while synaptic time constants $\tau_{e2}$ and $\tau_{i2}$ surpass $r > 0.92$. Synaptic coupling parameters $C_3$ and $C_4$ maintain $r \approx 0.85$, whereas sigmoid-related parameters ($\theta$, $\beta$, $r_{\max}$) exhibit comparatively lower identifiability ($r \approx 0.69\text{--}0.72$). Spatial consistency analyses (Figure~\ref{fig:sim_validation_all}\textbf{b}) reveal structured cortical heterogeneity: $\tau_{e2}$ achieves the highest average correlation across regions (mean $r = 0.87$), while inhibitory time constants display wider spatial variance.

\textbf{Inversion Degradation under Noise Perturbations:} Multi-level SNR stress tests demonstrate clear noise-type and parameter-specific dependencies (Figure~\ref{fig:sim_validation_all}\textbf{c}). At SNR $=30$\,dB, performance under EOG-like noise remains near clean baseline ($r = 0.81$ vs. $0.83$), whereas pink noise, white noise, and EMG perturbations drop to $r = 0.58$, $0.54$, and $0.53$, respectively. At SNR $=-5$\,dB, mean correlations decrease to $0.28$ (EOG), $0.24$ (EMG), $0.10$ (white noise), and $0.07$ (pink noise), though 40 of 48 parameter--noise combinations maintain positive correlations. Crucially, $\mathrm{delay\_scale}$ proves most resilient, decreasing from $r = 0.89$ (30 dB) to $r = 0.55$ ($-5$ dB). Negative correlations emerge only at low SNR ($\leq 5$\,dB) in localized coupling parameters ($C_1$--$C_3$), demarcating the performance under the evaluated surrogate-noise conditions of the model against broadband artifacts.

\textbf{Real EEG Closed-Loop Spatiotemporal Consistency:} In real clinical EEG, closed-loop reconstruction demonstrates moderate spectral and phase consistency. Temporally (Figure~\ref{fig:sim_validation_all}\textbf{d}), median PLV increases progressively from $\delta$ ($0.367$) and $\theta$ ($0.521$) to $\alpha$ ($0.679$). This band-ascending hierarchy matches the physiological dominance of $\theta/\alpha$ rhythms in resting-state dynamics, confirming that inferred biophysical parameters effectively retain instantaneous phase coupling and spectral topographies in closed-loop generation. Spectrally (Figure~\ref{fig:sim_validation_all}\textbf{e}), 1--40 Hz log PSD exhibits moderate-to-high linear consistency (channel mean $r = 0.674$), peaking in parieto-occipital and central channels (PZ: $0.734$, CZ: $0.732$). Resting-state $\alpha$-peak frequency absolute error averages $0.500$ Hz (FZ: $0.449$ Hz), while $1/f$ aperiodic slope error averages $1.566$. 

\subsection{Downstream Benchmark Performance}
\label{sec:downstream_eval}
Quantitative evaluations on four clinical benchmarks are summarized in Table~\ref{tab:main_results}. Overall, NeuroDyn-EEG demonstrates competitive, task-dependent advantages. On \textit{AD65}, NeuroDyn-EEG achieves the highest Balanced Accuracy (BACC, \textbf{0.829}) alongside the second-highest AUCROC and AUCPR. On \textit{PD31} and \textit{MDD}, NeuroDyn-EEG outperforms all compared models across all three evaluation metrics (BACC, AUCROC, AUCPR)(Supplementary Section~\ref{sec:supp_eval}).

On \textit{TUAB}, the model exhibits moderate performance. This outcome is expected given that the TUAB benchmark focuses on distinguishing normal physiological activity from general clinical abnormalities (heavily dominated by non-physiological artifacts such as muscle activity and electrode drift). Because NeuroDyn-EEG is rigorously constrained by generative biophysical equations, its architecture naturally filters non-physiological broadband signals, limiting its capacity as a generic artifact detector while enhancing its sensitivity to genuine neuropathological dynamics in tasks like \textit{AD65} and \textit{MDD}.

Furthermore, NeuroDyn-EEG demonstrates exceptional parameter efficiency (Table~\ref{tab:main_results}, \textit{Size} column): with only \textbf{2.43M} parameters, it surpasses substantially larger foundation models (e.g., the 33M \textit{BrainOmni base} and 5.8M \textit{LaBraM}) on multiple benchmarks.

\begin{table}[t]
\caption{Quantitative performance and model efficiency comparison. (a) Neurodegenerative disease datasets; (b) General clinical and psychiatric datasets. \textbf{Size} denotes the number of trainable parameters. \textbf{Bold} and \underline{underline} denote best and second-best results, respectively.}
\label{tab:main_results}
\centering
\small
\setlength{\tabcolsep}{3pt}
\begin{tabular}{l|c|ccc|ccc}
\toprule
\multirow{2}{*}{\textbf{Method}} & \multirow{2}{*}{\textbf{Model Size}} & \multicolumn{3}{c|}{\textbf{AD65}} & \multicolumn{3}{c}{\textbf{PD31}} \\
\cmidrule(lr){3-5} \cmidrule(lr){6-8}
 & & BACC & AUCROC & AUCPR & BACC & AUCROC & AUCPR \\
\midrule
LaBraM         & 5.8M  & 0.775$\pm$0.083 & 0.890$\pm$0.085 & 0.915$\pm$0.065 & 0.567$\pm$0.091 & 0.722$\pm$0.111 & 0.721$\pm$0.162 \\
CBraMod        & 4.9M  & 0.526$\pm$0.031 & 0.681$\pm$0.104 & 0.769$\pm$0.046 & 0.533$\pm$0.075 & 0.678$\pm$0.127 & 0.709$\pm$0.143 \\
BrainOmni tiny & 8.4M  & 0.767$\pm$0.140 & 0.853$\pm$0.121 & 0.896$\pm$0.089 & 0.533$\pm$0.075 & 0.756$\pm$0.214 & \underline{0.778$\pm$0.227} \\
BrainOmni base & 33M   & 0.713$\pm$0.116 & 0.871$\pm$0.103 & 0.917$\pm$0.065 & 0.600$\pm$0.149 & 0.683$\pm$0.239 & 0.722$\pm$0.213 \\
VAEEG          & 0.46M & \underline{0.798$\pm$0.136} & \textbf{0.938$\pm$0.071} & \textbf{0.954$\pm$0.050} & \underline{0.717$\pm$0.139} & \underline{0.789$\pm$0.159} & 0.751$\pm$0.191 \\
\midrule
\textbf{NeuroDyn-EEG} & 2.43M & \textbf{0.829$\pm$0.130} & \underline{0.910$\pm$0.124} & \underline{0.934$\pm$0.080} & \textbf{0.733$\pm$0.149} & \textbf{0.822$\pm$0.127} & \textbf{0.863$\pm$0.105} \\
\bottomrule
\end{tabular}

\vspace{1.5em} 

\begin{tabular}{l|c|ccc|ccc}
\toprule
\multirow{2}{*}{\textbf{Method}} & \multirow{2}{*}{\textbf{Model Size}} & \multicolumn{3}{c|}{\textbf{TUAB}} & \multicolumn{3}{c}{\textbf{MDD}} \\
\cmidrule(lr){3-5} \cmidrule(lr){6-8}
 & & BACC & AUCROC & AUCPR & BACC & AUCROC & AUCPR \\
\midrule
LaBraM         & 5.8M  & \underline{0.812$\pm$0.012} & \underline{0.865$\pm$0.013} & \underline{0.888$\pm$0.010} & 0.847$\pm$0.139 & 0.905$\pm$0.155 & 0.921$\pm$0.125 \\
CBraMod        & 4.9M  & 0.747$\pm$0.022 & 0.828$\pm$0.027 & 0.854$\pm$0.018 & 0.833$\pm$0.168 & 0.860$\pm$0.134 & 0.848$\pm$0.153 \\
BrainOmni tiny & 8.4M  & 0.779$\pm$0.017 & 0.834$\pm$0.016 & 0.867$\pm$0.008 & 0.827$\pm$0.130 & \underline{0.959$\pm$0.072} & 0.965$\pm$0.060 \\
BrainOmni base & 33M   & 0.788$\pm$0.006 & 0.845$\pm$0.017 & 0.875$\pm$0.008 & 0.843$\pm$0.052 & 0.957$\pm$0.061 & \underline{0.969$\pm$0.044} \\
VAEEG          & 0.46M & \textbf{0.837$\pm$0.016} & \textbf{0.922$\pm$0.013} & \textbf{0.910$\pm$0.015} & \underline{0.853$\pm$0.137} & 0.895$\pm$0.113 & 0.910$\pm$0.102 \\
\midrule
\textbf{NeuroDyn-EEG} & 2.43M & 0.726$\pm$0.011 & 0.816$\pm$0.028 & 0.834$\pm$0.015 & \textbf{0.927$\pm$0.077} & \textbf{0.980$\pm$0.045} & \textbf{0.984$\pm$0.036} \\
\bottomrule
\end{tabular}
\end{table}

\subsection{Case--Control AAL Patterns in Inferred JR Parameters}
\label{sec:jr_param_stats_results}

\begin{figure}[t]
    \centering
    \includegraphics[width=\textwidth]{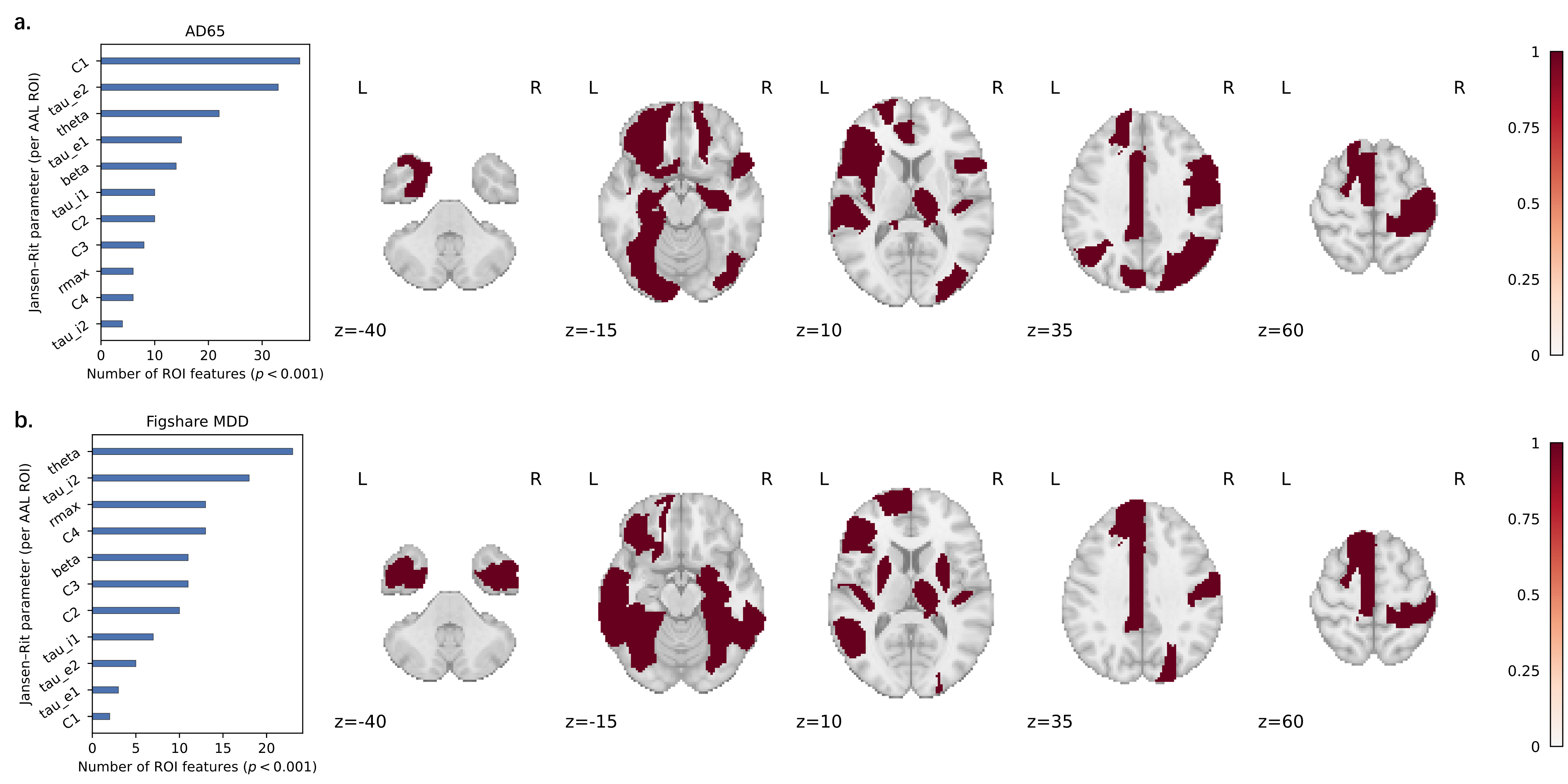}
    \caption{\textbf{Region-wise case--control structure of inferred JR parameters in AD65 and Figshare MDD.} (\textbf{a}) \textit{AD65}: Bar plot shows the number of AAL regions satisfying Benjamini--Hochberg corrected $q_{\mathrm{BH}}<0.001$ for each JR parameter family; right surface map displays the significance distribution for the top-ranked parameter \texttt{C1} across AAL-90 space. (\textbf{b}) \textit{Figshare MDD}: Following the same protocol, the top-ranked parameter is \texttt{$\theta$}. Colorbars represent binary significance masks (1 = region significant at $q_{\mathrm{BH}}<0.001$; 0 = non-significant). The global scalar $\mathrm{delay\_scale}$ is excluded from regional counts.}
    \label{fig:jr_downstream_param}
\end{figure}

Beyond classification metrics, we examine whether inferred biophysical parameters carry anatomically localized case--control information (Figure~\ref{fig:jr_downstream_param}). In \textit{AD65}, the parameter family exhibiting the most significant regions is the local synaptic connectivity parameter \texttt{$C_1$} (37/90 regions), followed by \texttt{$\tau_{e2}$} (33), \texttt{$\theta$} (22), \texttt{$\tau_{e1}$} (15), and \texttt{$\beta$} (14). Significant regions for \texttt{$C_1$} exhibit a distributed cross-system topography involving prefrontal and cingulate cortices, medial temporal structures, sensorimotor-parietal areas, thalamo-basal ganglia nodes, and occipitotemporal visual regions.

In \textit{Figshare MDD}, the top-ranked parameter is the neural population firing threshold \texttt{$\theta$} (23/90 regions), followed by \texttt{$\tau_{i2}$} (18), \texttt{$C_4$} and \texttt{$r_{max}$} (13 each), and \texttt{$C_3$} and \texttt{$\beta$} (11 each). Significant regions for \texttt{$\theta$} span prefrontal--cingulate, limbic--medial temporal, sensorimotor, temporo-occipital, and striato-thalamic networks. Complete statistical summaries are provided in Supplementary Tables~\ref{tab:supp_ad65_c1_stats} and \ref{tab:supp_mdd_theta_stats}.

The divergence between the two cohorts---AD65 dominated by local excitatory connectivity \texttt{$C_1$} and MDD dominated by population threshold \texttt{$\theta$}---highlights distinct cohort-level parameter–region patterns in the inferred parameter space, providing testable hypotheses for downstream mechanistic investigations.

\section{Discussion}
This study introduces and evaluates NeuroDyn-EEG, an EEG pre-training framework that integrates neurodynamical generative priors. Unlike conventional approaches that directly optimize for classification labels or extract unconstrained latent embeddings, NeuroDyn-EEG deterministically maps each EEG segment into a Jansen--Rit (JR) parameter field parcellated across the AAL-90 atlas. Grounded in network neuroscience principles, this formulation enables rigorous case--control hypothesis testing across distributed regional parameter profiles beyond standard classification metrics \citep{bassett2017network,fornito2015connectomics}.

Across simulation-domain benchmarks and empirical closed-loop reconstructions, the framework demonstrated robust parameter recoverability and observational fidelity. Surrogate noise experiments revealed selective perturbation robustness; notably, $\mathrm{delay\_scale}$ maintained high stability across all four surrogate noise types and SNR regimes, indicating that the model reliably preserves recoverable information pertinent to global propagation delays even from partially corrupted scalp recordings. In empirical resting-state EEG, the inverse-to-forward closed-loop reconstruction yielded Phase-Locking Values (PLV) of $0.521$ and $0.679$ in the $\theta$ and $\alpha$ bands, respectively \citep{lachaux1999measuring}. Furthermore, the spectral consistency observed across parieto-occipital and central montages qualitatively matched the canonical scalp topography of resting-state $\alpha$ rhythms. These findings demonstrate that the inferred biophysical parameters effectively retain macroscopic oscillatory dynamics within the generative loop, thereby enhancing the empirical testability of the derived representations.

On downstream clinical evaluations, NeuroDyn-EEG demonstrated task-dependent competitive advantages across heterogeneous cohorts, achieving superior performance across all or subset metrics on AD65, PD31, and MDD, while exhibiting suboptimal performance on TUAB. This discrepancy arises because TUAB focuses on broad clinical normal-versus-abnormal screening \citep{obeid2016tuh}, which diverges fundamentally from phenotype-specific case--control paradigms. Compared to purely data-driven baselines reliant on massive corpora (e.g., masked reconstruction architectures such as LaBraM and BrainOmni), the primary computational merit of NeuroDyn-EEG lies in anatomically indexed and biophysically defined representation. In computational psychiatry and neurology, raw classification accuracy alone frequently fails to satisfy clinical demands for transparency and mechanistic interpretability \citep{rudin2019stop}. By integrating the non-linear representation capacity of deep architectures with the biophysical constraints of the Jansen--Rit neural mass model, our framework not only mitigates overfitting risks common in small-sample clinical cohorts, but critically establishes a direct computational pipeline bridging phenomenological scalp potentials to generative synaptic and dynamical mechanisms. This synergy between representation learning and simulation-based inference (SBI) \citep{goncalves2020training} significantly expands the analytical scope of electrophysiological feature engineering.

Region-wise parameter statistics derived from NeuroDyn-EEG provide well-defined computational targets for disease-specific network modeling; moreover, cross-cohort parameter divergence offers candidate mechanistic hypotheses into distinct pathways of systemic decompensation. In the AD65 cohort, case-control differences primarily indexed by the local excitatory connectivity parameter \texttt{C1} spanned the default mode network (DMN), limbic memory systems, and cortico-thalamic/basal ganglia circuits. These alterations are consistent with documented structural degeneration, functional desynchronization, and EEG source connectivity disruptions \citep{jeong2004eeg,stam2005eeg,cassani2018systematic,palop2016network,brier2012loss,hata2016functional,buckner2005molecular,seeley2009neurodegenerative}, lending dynamical support to the characterization of Alzheimer's disease as a large-scale network "disconnection syndrome" \citep{delbeuck2003alzheimer,palesi2016exploring}. Conversely, in the Figshare MDD cohort, the neural population sigmoidal firing threshold \texttt{$\theta$} emerged as the principal differentiating parameter. Regions exhibiting significant alterations encompassed emotion-regulation networks, cortico-limbic circuits, and cortico-striatal-thalamic loops—concordant with structural and functional alterations reported in large-scale MDD cohorts \citep{drevets2008brain,price2012neural,kaiser2015large,mulders2015resting,menon2011large,javaheripour2021altered}—while highlighting underlying systemic excitation/inhibition (E/I) imbalances and aberrant neuronal firing gains \citep{hu2023brain,uhlhaas2012neuronal,sanacora2012towards}. These anatomically grounded findings demonstrate the efficacy of profiling heterogeneous neuropsychiatric disorders within a unified dynamical parameter space, bridging representation learning with mechanistic hypothesis generation.

Despite these promising results, several methodological boundaries warrant cautious interpretation: First, regarding parameter identifiability and physiological correspondence, observational fidelity in closed-loop reconstruction does not guarantee unique mathematical identifiability. The inferred JR parameters represent constrained point estimates capturing macroscale shifts in firing thresholds or effective coupling; they should not be conflated with direct physical measurements of microscale synaptic currents or neurotransmitter concentrations (e.g., GABA). Furthermore, the potential influence of discriminative inductive biases introduced during supervised fine-tuning on the final parameter space cannot be fully ruled out. Second, regarding experimental scope, several underlying mechanisms require further elucidation. For instance, the differential parametric sensitivity observed across distinct surrogate noise regimes (e.g., additive white Gaussian noise versus $1/f$ noise) warrants systematic ablation studies. Similarly, the moderate performance observed on the broad anomaly detection task (TUAB) requires targeted error analysis to decouple task-definition mismatches from model-specific inductive biases. Finally, regarding interpretive scope and spatial resolution, the current post hoc analyses capture cohort-level regional parameter shifts and cannot independently establish cross-modal causal pathways. Additionally, the AAL-90 template operates at a relatively coarse macroscopic resolution, which may obscure fine-grained localized dynamical anomalies. Future investigations should incorporate larger-scale, longitudinal, and multimodal neuroimaging constraints (e.g., structural MRI, DTI, and PET) to validate whether key dynamical parameters, such as \texttt{$C_1$} and \texttt{$\theta$}, can serve as robust and clinically translatable computational biomarkers.

\section{Conclusion}
\label{sec:conclusion}
We developed NeuroDyn-EEG, an EEG modeling framework constrained by neural dynamics that combines an extended Jansen-Rit generator, leadfield-based forward projection, and deterministic point estimation inspired by simulation-based inference. Rather than learning discriminative representations without explicit anatomical indexing, the framework maps each EEG segment to a field of JR parameters indexed by AAL-90 regions, enabling parameter-level group comparisons in addition to downstream classification.

We evaluated this objective at three levels. First, in controlled simulations, NeuroDyn-EEG recovered spatially heterogeneous parameters with parameter-dependent fidelity and showed varying degrees of degradation under different surrogate-noise conditions. Second, in real resting-state EEG, waveforms reconstructed by the same forward model preserved aspects of spectral shape, dominant rhythms, and phase organization. These findings support model-internal consistency between the inferred parameters and macroscopic rhythmic structure, but they do not establish unique identifiability or direct physiological validity. Third, across the heterogeneous AD65, PD31, Figshare MDD, and TUAB cohorts, the framework achieved competitive but dataset-dependent downstream performance with a compact parameter count.

The parameter-level outputs provide testable candidate hypotheses about disease-associated dynamics. In AD65, regional differences were led by $C_1$, representing local connectivity from pyramidal cells to excitatory interneurons/stellate cells, and involved prefrontal, cingulate, medial temporal, parietal/sensorimotor, thalamic/basal-ganglia, and occipitotemporal regions. In Figshare MDD, the leading parameter was the neural-population firing threshold $\theta$, with differences involving prefrontal-limbic-temporal, occipital, sensorimotor, striatal, and thalamic regions. The two datasets therefore showed different parameter rankings; however, these differences may also reflect cohort composition, signal quality, and supervised training.

Overall, NeuroDyn-EEG combines predictive utility with anatomically indexed, inspectable model variables in the current benchmarks. By producing region-resolved parameters that can be tested statistically, it offers a computational link between EEG representation learning and mechanistic hypothesis generation. Validation in larger multicentre, longitudinal, and treatment-response cohorts, together with MRI, diffusion MRI, PET, or neurotransmitter-related measurements, will be necessary to determine whether parameters such as $C_1$ and $\theta$ are sufficiently stable for disease stratification, prognosis, or prediction of treatment response.

\bibliographystyle{unsrtnat}
\bibliography{references}

@article{jeong2004eeg,
  author     = {Jeong, J.},
  title      = {{EEG} dynamics in patients with {Alzheimer's} disease},
  journal    = {Clinical Neurophysiology},
  year       = {2004},
  volume     = {115},
  number     = {7},
  pages      = {1490--1505},
  doi        = {10.1016/j.clinph.2004.01.001},
}

@article{stam2005eeg,
  author     = {Stam, C. J. and van der Made, Y. and Pijnenburg, Y. A. L. and others},
  title      = {{EEG} synchronization in mild cognitive impairment and {Alzheimer's} disease},
  journal    = {Acta Neurologica Scandinavica},
  year       = {2003},
  volume     = {108},
  number     = {2},
  pages      = {90--96},
  doi        = {10.1034/j.1600-0404.2003.02067.x},
}

@article{palop2016network,
  author     = {Palop, J. J. and Mucke, L.},
  title      = {Network abnormalities and interneuron dysfunction in {Alzheimer} disease},
  journal    = {Nature Reviews Neuroscience},
  year       = {2016},
  volume     = {17},
  number     = {12},
  pages      = {777--792},
  doi        = {10.1038/nrn.2016.141},
}

@article{kaiser2015large,
  author     = {Kaiser, R. H. and Andrews-Hanna, J. R. and Wager, T. D. and others},
  title      = {Large-scale network dysfunction in major depressive disorder: a meta-analysis of resting-state functional connectivity},
  journal    = {JAMA Psychiatry},
  year       = {2015},
  volume     = {72},
  number     = {6},
  pages      = {603--611},
  doi        = {10.1001/jamapsychiatry.2015.0071},
}

@article{mulders2015resting,
  author     = {Mulders, P. C. and van Eijndhoven, P. F. and Schene, A. H. and others},
  title      = {Resting-state functional connectivity in major depressive disorder: a review},
  journal    = {Neuroscience \& Biobehavioral Reviews},
  year       = {2015},
  volume     = {56},
  pages      = {330--344},
  doi        = {10.1016/j.neubiorev.2015.07.014},
}

@article{sanacora1999reduced,
  author     = {Sanacora, G. and Mason, G. F. and Rothman, D. L. and others},
  title      = {Reduced cortical gamma-aminobutyric acid levels in depressed patients determined by proton magnetic resonance spectroscopy},
  journal    = {Archives of General Psychiatry},
  year       = {1999},
  volume     = {56},
  number     = {11},
  pages      = {1043--1047},
  doi        = {10.1001/archpsyc.56.11.1043},
}

@article{jansen1995electroencephalogram,
  author     = {Jansen, B. H. and Rit, V. G.},
  title      = {Electroencephalogram and visual evoked potential generation in a mathematical model of coupled cortical columns},
  journal    = {Biological Cybernetics},
  year       = {1995},
  volume     = {73},
  number     = {4},
  pages      = {357--366},
  doi        = {10.1007/bf00199471},
}

@article{david2003modeling,
  author     = {David, O. and Friston, K. J.},
  title      = {A neural mass model for {MEG}/{EEG}: coupling and neuronal dynamics},
  journal    = {NeuroImage},
  year       = {2003},
  volume     = {20},
  number     = {3},
  pages      = {1743--1755},
  doi        = {10.1016/j.neuroimage.2003.07.015},
}

@article{cranmer2020frontier,
  author     = {Cranmer, K. and Brehmer, J. and Louppe, G.},
  title      = {The frontier of simulation-based inference},
  journal    = {Proceedings of the National Academy of Sciences},
  year       = {2020},
  volume     = {117},
  number     = {48},
  pages      = {30055--30062},
  doi        = {10.1073/pnas.1912789117},
}

@article{gramfort2014mne,
  author     = {Gramfort, A. and Luessi, M. and Larson, E. and others},
  title      = {{MNE} software for processing {MEG} and {EEG} data},
  journal    = {NeuroImage},
  year       = {2014},
  volume     = {86},
  pages      = {446--460},
  doi        = {10.1016/j.neuroimage.2013.10.027},
}

@article{tzourio2002automated,
  author     = {Tzourio-Mazoyer, N. and Landeau, B. and Papathanassiou, D. and others},
  title      = {Automated anatomical labeling of activations in {SPM} using a macroscopic anatomical parcellation of the {MNI} {MRI} single-subject brain},
  journal    = {NeuroImage},
  year       = {2002},
  volume     = {15},
  number     = {1},
  pages      = {273--289},
  doi        = {10.1006/nimg.2001.0978},
}

@article{jasper1958ten,
  author     = {Jasper, H. H.},
  title      = {The ten-twenty electrode system of the {International Federation}},
  journal    = {Electroencephalography and Clinical Neurophysiology},
  year       = {1958},
  volume     = {10},
  pages      = {371--375},
  url        = {https://asenic.ru/fishka/austral/jasper1957.pdf},
}

@inproceedings{ioffe2015batch,
  author     = {Ioffe, S. and Szegedy, C.},
  title      = {Batch normalization: accelerating deep network training by reducing internal covariate shift},
  booktitle  = {Proceedings of the 32nd International Conference on Machine Learning},
  year       = {2015},
  volume     = {37},
  pages      = {448--456},
  series     = {Proceedings of Machine Learning Research},
  publisher  = {PMLR},
  url        = {https://proceedings.mlr.press/v37/ioffe15.html},
}

@inproceedings{maas2013rectifier,
  author     = {Maas, A. L. and Hannun, A. Y. and Ng, A. Y.},
  title      = {Rectifier nonlinearities improve neural network acoustic models},
  booktitle  = {ICML Workshop on Deep Learning for Audio, Speech and Language Processing},
  year       = {2013},
  url        = {https://ai.stanford.edu/~amaas/papers/relu_hybrid_icml2013_final.pdf},
}

@article{welch1947generalization,
  author     = {Welch, B. L.},
  title      = {The generalization of `{Student's}' problem when several different population variances are involved},
  journal    = {Biometrika},
  year       = {1947},
  volume     = {34},
  number     = {1-2},
  pages      = {28--35},
  doi        = {10.1093/biomet/34.1-2.28},
}

@article{benjamini1995controlling,
  author     = {Benjamini, Y. and Hochberg, Y.},
  title      = {Controlling the false discovery rate: a practical and powerful approach to multiple testing},
  journal    = {Journal of the Royal Statistical Society: Series B (Methodological)},
  year       = {1995},
  volume     = {57},
  number     = {1},
  pages      = {289--300},
  doi        = {10.1111/j.2517-6161.1995.tb02031.x},
}

@inproceedings{jiang2024labram,
  author     = {Jiang, W.-B. and Zhao, L.-M. and Lu, B.-L.},
  title      = {Large brain model for learning generic representations with tremendous {EEG} data in {BCI}},
  booktitle  = {The Twelfth International Conference on Learning Representations},
  year       = {2024},
  url        = {https://openreview.net/forum?id=QzTpTRVtrP},
}

@inproceedings{wang2025cbramod,
  author     = {Wang, J. and Zhao, S. and Luo, Z. and others},
  title      = {{CBraMod}: a criss-cross brain foundation model for {EEG} decoding},
  booktitle  = {The Thirteenth International Conference on Learning Representations},
  year       = {2025},
  url        = {https://openreview.net/forum?id=NPNUHgHF2w},
}

@inproceedings{yang2024brainomni,
  author     = {Xiao, Q. and Cui, Z. and Zhang, C. and others},
  title      = {{BrainOmni}: a brain foundation model for unified {EEG} and {MEG} signals},
  booktitle  = {Advances in Neural Information Processing Systems},
  year       = {2025},
  volume     = {38},
  pages      = {46081--46114},
  publisher  = {Curran Associates, Inc.},
  doi        = {10.52202/085713-1375},
  url        = {https://papers.neurips.cc/paper_files/paper/2025/hash/3aef4a18c2646b9d11d9ab4d9bec72c8-Abstract-Conference.html},
}

@article{zhang2023vaeeg,
  author     = {Zhao, T. and Cui, Y. and Ji, T. and others},
  title      = {{VAEEG}: variational auto-encoder for extracting {EEG} representation},
  journal    = {NeuroImage},
  year       = {2024},
  volume     = {304},
  pages      = {120946},
  doi        = {10.1016/j.neuroimage.2024.120946},
}

@article{bassett2017network,
  author     = {Bassett, D. S. and Sporns, O.},
  title      = {Network neuroscience},
  journal    = {Nature Neuroscience},
  year       = {2017},
  volume     = {20},
  number     = {3},
  pages      = {353--364},
  doi        = {10.1038/nn.4502},
}

@article{fornito2015connectomics,
  author     = {Fornito, A. and Zalesky, A. and Breakspear, M.},
  title      = {The connectomics of brain disorders},
  journal    = {Nature Reviews Neuroscience},
  year       = {2015},
  volume     = {16},
  number     = {3},
  pages      = {159--172},
  doi        = {10.1038/nrn3901},
}

@article{deco2011emerging,
  author     = {Deco, G. and Jirsa, V. K. and McIntosh, A. R.},
  title      = {Emerging concepts for the dynamical organization of resting-state activity in the brain},
  journal    = {Nature Reviews Neuroscience},
  year       = {2011},
  volume     = {12},
  number     = {1},
  pages      = {43--56},
  doi        = {10.1038/nrn2961},
}

@article{sanzleon2013virtualbrain,
  author     = {Sanz Leon, P. and Knock, S. A. and Woodman, M. M. and others},
  title      = {{The Virtual Brain}: a simulator of primate brain network dynamics},
  journal    = {Frontiers in Neuroinformatics},
  year       = {2013},
  volume     = {7},
  pages      = {10},
  doi        = {10.3389/fninf.2013.00010},
}

@article{brier2012loss,
  author     = {Brier, M. R. and Thomas, J. B. and Snyder, A. Z. and others},
  title      = {Loss of intranetwork and internetwork resting state functional connections with {Alzheimer's} disease progression},
  journal    = {Journal of Neuroscience},
  year       = {2012},
  volume     = {32},
  number     = {26},
  pages      = {8890--8899},
  doi        = {10.1523/jneurosci.5698-11.2012},
}

@article{buckner2005molecular,
  author     = {Buckner, R. L. and Snyder, A. Z. and Shannon, B. J. and others},
  title      = {Molecular, structural, and functional characterization of {Alzheimer's} disease: evidence for a relationship between default activity, amyloid, and memory},
  journal    = {Journal of Neuroscience},
  year       = {2005},
  volume     = {25},
  number     = {34},
  pages      = {7709--7717},
  doi        = {10.1523/jneurosci.2177-05.2005},
}

@article{seeley2009neurodegenerative,
  author     = {Seeley, W. W. and Crawford, R. K. and Zhou, J. and others},
  title      = {Neurodegenerative diseases target large-scale human brain networks},
  journal    = {Neuron},
  year       = {2009},
  volume     = {62},
  number     = {1},
  pages      = {42--52},
  doi        = {10.1016/j.neuron.2009.03.024},
}

@article{drevets2008brain,
  author     = {Drevets, W. C. and Price, J. L. and Furey, M. L.},
  title      = {Brain structural and functional abnormalities in mood disorders: implications for neurocircuitry models of depression},
  journal    = {Brain Structure and Function},
  year       = {2008},
  volume     = {213},
  number     = {1-2},
  pages      = {93--118},
  doi        = {10.1007/s00429-008-0189-x},
}

@article{price2012neural,
  author     = {Price, J. L. and Drevets, W. C.},
  title      = {Neural circuits underlying the pathophysiology of mood disorders},
  journal    = {Trends in Cognitive Sciences},
  year       = {2012},
  volume     = {16},
  number     = {1},
  pages      = {61--71},
  doi        = {10.1016/j.tics.2011.12.011},
}

@article{menon2011large,
  author     = {Menon, V.},
  title      = {Large-scale brain networks and psychopathology: a unifying triple network model},
  journal    = {Trends in Cognitive Sciences},
  year       = {2011},
  volume     = {15},
  number     = {10},
  pages      = {483--506},
  doi        = {10.1016/j.tics.2011.08.003},
}

@article{obeid2016tuh,
  author     = {Obeid, I. and Picone, J.},
  title      = {The temple university hospital {EEG} data corpus},
  journal    = {Frontiers in Neuroscience},
  year       = {2016},
  volume     = {10},
  pages      = {196},
  doi        = {10.3389/fnins.2016.00196},
}

@article{schirrmeister2017deep,
  author     = {Schirrmeister, R. T. and Springenberg, J. T. and Fiederer, L. D. J. and others},
  title      = {Deep learning with convolutional neural networks for {EEG} decoding and visualization},
  journal    = {Human Brain Mapping},
  year       = {2017},
  volume     = {38},
  number     = {11},
  pages      = {5391--5420},
  doi        = {10.1002/hbm.23730},
}

@article{lawhern2018eegnet,
  author     = {Lawhern, V. J. and Solon, A. J. and Waytowich, N. R. and others},
  title      = {{EEGNet}: a compact convolutional neural network for {EEG}-based brain--computer interfaces},
  journal    = {Journal of Neural Engineering},
  year       = {2018},
  volume     = {15},
  number     = {5},
  pages      = {056013},
  doi        = {10.1088/1741-2552/aace8c},
}

@article{roy2019deep,
  author     = {Roy, Y. and Banville, H. and Albuquerque, I. and others},
  title      = {Deep learning-based electroencephalography analysis: a systematic review},
  journal    = {Journal of Neural Engineering},
  year       = {2019},
  volume     = {16},
  number     = {5},
  pages      = {051001},
  doi        = {10.1088/1741-2552/ab260c},
}

@book{nunez2006electric,
  author     = {Nunez, P. L. and Srinivasan, R.},
  title      = {Electric fields of the brain: the neurophysics of {EEG}},
  year       = {2006},
  edition    = {2},
  publisher  = {Oxford University Press},
  doi        = {10.1093/acprof:oso/9780195050387.001.0001},
}

@article{buzsaki2012origin,
  author     = {Buzs{\'{a}}ki, G. and Anastassiou, C. A. and Koch, C.},
  title      = {The origin of extracellular fields and currents --- {EEG}, {ECoG}, {LFP} and spikes},
  journal    = {Nature Reviews Neuroscience},
  year       = {2012},
  volume     = {13},
  number     = {6},
  pages      = {407--420},
  doi        = {10.1038/nrn3241},
}

@article{grech2008review,
  author     = {Grech, R. and Cassar, T. and Muscat, J. and others},
  title      = {Review on solving the inverse problem in {EEG} source analysis},
  journal    = {Journal of NeuroEngineering and Rehabilitation},
  year       = {2008},
  volume     = {5},
  number     = {1},
  pages      = {25},
  doi        = {10.1186/1743-0003-5-25},
}

@article{michel2019eegsource,
  author     = {Michel, C. M. and Brunet, D.},
  title      = {{EEG} source imaging: a practical review of the analysis steps},
  journal    = {Frontiers in Neurology},
  year       = {2019},
  volume     = {10},
  pages      = {325},
  doi        = {10.3389/fneur.2019.00325},
}

@article{goncalves2020training,
  author     = {Gon{\c{c}}alves, P. J. and Lueckmann, J.-M. and Deistler, M. and others},
  title      = {Training deep neural density estimators to identify mechanistic models of neural dynamics},
  journal    = {eLife},
  year       = {2020},
  volume     = {9},
  pages      = {e56261},
  doi        = {10.7554/elife.56261},
}

@article{tejero2020sbi,
  author     = {Tejero-Cantero, A. and Boelts, J. and Deistler, M. and others},
  title      = {{sbi}: a toolkit for simulation-based inference},
  journal    = {Journal of Open Source Software},
  year       = {2020},
  volume     = {5},
  number     = {52},
  pages      = {2505},
  doi        = {10.21105/joss.02505},
}

@article{boelts2024sbireloaded,
  author     = {Boelts, J. and Deistler, M. and Gloeckler, M. and others},
  title      = {{sbi} reloaded: a toolkit for simulation-based inference workflows},
  journal    = {Journal of Open Source Software},
  year       = {2025},
  volume     = {10},
  number     = {108},
  pages      = {7754},
  doi        = {10.21105/joss.07754},
}

@article{lachaux1999measuring,
  author     = {Lachaux, J.-P. and Rodriguez, E. and Martinerie, J. and others},
  title      = {Measuring phase synchrony in brain signals},
  journal    = {Human Brain Mapping},
  year       = {1999},
  volume     = {8},
  number     = {4},
  pages      = {194--208},
  doi        = {10.1002/(sici)1097-0193(1999)8:4<194::aid-hbm4>3.0.co;2-c},
}

@article{donoghue2020parameterizing,
  author     = {Donoghue, T. and Haller, M. and Peterson, E. J. and others},
  title      = {Parameterizing neural power spectra into periodic and aperiodic components},
  journal    = {Nature Neuroscience},
  year       = {2020},
  volume     = {23},
  number     = {12},
  pages      = {1655--1665},
  doi        = {10.1038/s41593-020-00744-x},
}

@article{cassani2018systematic,
  author     = {Cassani, R. and Estarellas, M. and San-Martin, R. and others},
  title      = {Systematic review on resting-state {EEG} for {Alzheimer's} disease diagnosis and progression assessment},
  journal    = {Disease Markers},
  year       = {2018},
  volume     = {2018},
  pages      = {5174815},
  doi        = {10.1155/2018/5174815},
}

@book{bernardo1994bayesian,
  author     = {Bernardo, J. M. and Smith, A. F. M.},
  title      = {{Bayesian} theory},
  year       = {1994},
  publisher  = {John Wiley \& Sons},
  doi        = {10.1002/9780470316870},
}

@inproceedings{lueckmann2021benchmarking,
  author     = {Lueckmann, J.-M. and Boelts, J. and Greenberg, D. and others},
  title      = {Benchmarking simulation-based inference},
  booktitle  = {Proceedings of the 24th International Conference on Artificial Intelligence and Statistics},
  year       = {2021},
  volume     = {130},
  pages      = {343--351},
  series     = {Proceedings of Machine Learning Research},
  publisher  = {PMLR},
  url        = {https://proceedings.mlr.press/v130/lueckmann21a.html},
}

@article{talts2018sbc,
  author     = {Talts, S. and Betancourt, M. and Simpson, D. and others},
  title      = {Validating {Bayesian} inference algorithms with simulation-based calibration},
  journal    = {arXiv preprint arXiv:1804.06788},
  year       = {2018},
  doi        = {10.48550/arXiv.1804.06788},
  url        = {https://arxiv.org/abs/1804.06788},
}

@article{hata2016functional,
  author     = {Hata, M. and Kazui, H. and Tanaka, T. and others},
  title      = {Functional connectivity assessed by resting state {EEG} correlates with cognitive decline of {Alzheimer's} disease -- an {eLORETA} study},
  journal    = {Clinical Neurophysiology},
  year       = {2016},
  volume     = {127},
  number     = {2},
  pages      = {1269--1278},
  doi        = {10.1016/j.clinph.2015.10.030},
}

@article{sun2024seizure,
  author     = {Sun, R. and Sohrabpour, A. and Joseph, B. and others},
  title      = {Seizure sources can be imaged from scalp {EEG} by means of biophysically constrained deep neural networks},
  journal    = {Advanced Science},
  year       = {2024},
  volume     = {11},
  number     = {47},
  pages      = {2405246},
  doi        = {10.1002/advs.202405246},
}

@article{javaheripour2021altered,
  author     = {Javaheripour, N. and Li, M. and Chand, T. and others},
  title      = {Altered resting-state functional connectome in major depressive disorder: a mega-analysis from the {PsyMRI} consortium},
  journal    = {Translational Psychiatry},
  year       = {2021},
  volume     = {11},
  number     = {1},
  pages      = {511},
  doi        = {10.1038/s41398-021-01619-w},
}

@article{skoch2022human,
  author     = {{\v{S}}koch, A. and Reh{\'{a}}k Bu{\v{c}}kov{\'{a}}, B. and Mare{\v{s}}, J. and others},
  title      = {Human brain structural connectivity matrices--ready for modelling},
  journal    = {Scientific Data},
  year       = {2022},
  volume     = {9},
  number     = {1},
  pages      = {486},
  doi        = {10.1038/s41597-022-01596-9},
}

@article{delbeuck2003alzheimer,
  author     = {Delbeuck, X. and Van der Linden, M. and Collette, F.},
  title      = {{Alzheimer's} disease as a disconnection syndrome?},
  journal    = {Neuropsychology Review},
  year       = {2003},
  volume     = {13},
  number     = {2},
  pages      = {79--92},
  doi        = {10.1023/a:1023832305702},
}

@article{palesi2016exploring,
  author     = {Palesi, F. and Castellazzi, G. and Casiraghi, L. and others},
  title      = {Exploring patterns of alteration in {Alzheimer's} disease brain networks: a combined structural and functional connectomics analysis},
  journal    = {Frontiers in Neuroscience},
  year       = {2016},
  volume     = {10},
  pages      = {380},
  doi        = {10.3389/fnins.2016.00380},
}

@article{hu2023brain,
  author     = {Hu, Y.-T. and Tan, Z.-L. and Hirjak, D. and others},
  title      = {Brain-wide changes in excitation-inhibition balance of major depressive disorder: a systematic review of topographic patterns of {GABA}- and glutamatergic alterations},
  journal    = {Molecular Psychiatry},
  year       = {2023},
  volume     = {28},
  number     = {8},
  pages      = {3257--3266},
  doi        = {10.1038/s41380-023-02193-x},
}

@article{uhlhaas2012neuronal,
  author     = {Uhlhaas, P. J. and Singer, W.},
  title      = {Neuronal dynamics and neuropsychiatric disorders: toward a translational paradigm for dysfunctional large-scale networks},
  journal    = {Neuron},
  year       = {2012},
  volume     = {75},
  number     = {6},
  pages      = {963--980},
  doi        = {10.1016/j.neuron.2012.09.004},
}

@article{sanacora2012towards,
  author     = {Sanacora, G. and Treccani, G. and Popoli, M.},
  title      = {Towards a glutamate hypothesis of depression: an emerging frontier of neuropsychopharmacology for mood disorders},
  journal    = {Neuropharmacology},
  year       = {2012},
  volume     = {62},
  number     = {1},
  pages      = {63--77},
  doi        = {10.1016/j.neuropharm.2011.07.036},
}

@article{rudin2019stop,
  author     = {Rudin, C.},
  title      = {Stop explaining black box machine learning models for high stakes decisions and use interpretable models instead},
  journal    = {Nature Machine Intelligence},
  year       = {2019},
  volume     = {1},
  number     = {5},
  pages      = {206--215},
  doi        = {10.1038/s42256-019-0048-x},
}

@article{croft2000eog,
  author     = {Croft, R. J. and Barry, R. J.},
  title      = {Removal of ocular artifact from the {EEG}: a review},
  journal    = {Neurophysiologie Clinique/Clinical Neurophysiology},
  year       = {2000},
  volume     = {30},
  number     = {1},
  pages      = {5--19},
  doi        = {10.1016/s0987-7053(00)00055-1},
}

@article{muthukumaraswamy2013high,
  author     = {Muthukumaraswamy, S. D.},
  title      = {High-frequency brain activity and muscle artifacts in {MEG}/{EEG}: a review and recommendations},
  journal    = {Frontiers in Human Neuroscience},
  year       = {2013},
  volume     = {7},
  pages      = {138},
  doi        = {10.3389/fnhum.2013.00138},
}

@article{elbert1985removal,
  author     = {Elbert, T. and Lutzenberger, W. and Rockstroh, B. and others},
  title      = {Removal of ocular artifacts from the {EEG} --- a biophysical approach to the eog},
  journal    = {Electroencephalography and Clinical Neurophysiology},
  year       = {1985},
  volume     = {60},
  number     = {5},
  pages      = {455--463},
  doi        = {10.1016/0013-4694(85)91020-x},
}

@article{goncharova2003spectral,
  author     = {Goncharova, I. I. and McFarland, D. J. and Vaughan, T. M. and others},
  title      = {Emg contamination of {EEG}: spectral and topographical characteristics},
  journal    = {Clinical Neurophysiology},
  year       = {2003},
  volume     = {114},
  number     = {9},
  pages      = {1580--1593},
  doi        = {10.1016/s1388-2457(03)00093-2},
}

@article{he2014scale,
  author     = {He, B. J.},
  title      = {Scale-free brain activity: past, present, and future},
  journal    = {Trends in Cognitive Sciences},
  year       = {2014},
  volume     = {18},
  number     = {9},
  pages      = {480--487},
  doi        = {10.1016/j.tics.2014.04.003},
}

@article{li2006blink,
  author     = {Li, Y. and Ma, Z. and Lu, W. and others},
  title      = {Automatic removal of the eye blink artifact from {EEG} using an {ICA}-based template matching approach},
  journal    = {Physiological Measurement},
  year       = {2006},
  volume     = {27},
  number     = {4},
  pages      = {425--436},
  doi        = {10.1088/0967-3334/27/4/008},
}

@article{uriguen2015eeg,
  author     = {Urig{\"{u}}en, J. A. and Garcia-Zapirain, B.},
  title      = {{EEG} artifact removal---state-of-the-art and guidelines},
  journal    = {Journal of Neural Engineering},
  year       = {2015},
  volume     = {12},
  number     = {3},
  pages      = {031001},
  doi        = {10.1088/1741-2560/12/3/031001},
}

@article{zhang2021eegdenoisenet,
  author     = {Zhang, H. and Zhao, M. and Wei, C. and others},
  title      = {{EEGdenoiseNet}: a benchmark dataset for deep learning solutions of {EEG} denoising},
  journal    = {Journal of Neural Engineering},
  year       = {2021},
  volume     = {18},
  number     = {5},
  pages      = {056057},
  doi        = {10.1088/1741-2552/ac2bf8},
}

@article{coroneloliveros2026multifrequency,
  author     = {Coronel-Oliveros, C. and Lehue, F. and Herzog, R. and others},
  title      = {A multi-frequency whole-brain neural mass model with homeostatic feedback inhibition},
  journal    = {PLOS Computational Biology},
  year       = {2026},
  volume     = {22},
  number     = {5},
  pages      = {e1013463},
  doi        = {10.1371/journal.pcbi.1013463},
}

\end{document}


\renewcommand{\thefigure}{S\arabic{figure}}
\renewcommand{\thetable}{S\arabic{table}}
\renewcommand{\thealgorithm}{S\arabic{algorithm}}
\renewcommand{\tablename}{Supplementary Table}
\floatname{algorithm}{Supplementary Algorithm}

\maketitle

\paragraph{Overview}

This document compiles extended tables, the hierarchical, physiology-informed parameter sampler used to generate the training corpus (Section~S1), training implementation details (Section~S2), baseline-specific preprocessing specifications and NeuroDyn-EEG clinical cohort harmonization rules (Section~S3), downstream evaluation setups (Section~S4), simulation experiment metrics (Section~S5), surrogate noise synthesis for robustness experiments (Section~S6), clinical cohort JR parameter statistical tables and AAL naming mapping notes (Section~S7), as well as the training algorithm referenced in the main text. In the main text, a framework schematic summarizes the three-stage modeling and validation pipeline of NeuroDyn-EEG, while simulation validation and downstream/parameter-level analyses are reported in dedicated result figures and tables. Tables are numbered S1--S7; extended subsections are numbered S1--S7; the algorithm is designated as Supplementary Algorithm~S1.


\begin{table}[t]
\centering
\caption{Sampled dynamical parameters for forward simulation. Bounds specify the simulation prior.}
\label{tab:supp_sampling}
\small
\setlength{\tabcolsep}{4pt}
\begin{tabular}{@{}p{0.10\linewidth}p{0.385\linewidth}p{0.18\linewidth}p{0.09\linewidth}p{0.175\linewidth}@{}}
\toprule
\textbf{Parameter} & \textbf{Description} & \textbf{Range} & \textbf{Unit} & \textbf{Distribution} \\
\midrule
$\tau_{e1}$ & Excitatory time constant, slow branch & $[10,35]$ & $\mathrm{ms}$ & HTN \\
$\tau_{i1}$ & Inhibitory time constant, slow branch & $[10,35]$ & $\mathrm{ms}$ & Conditional uniform \\
$\tau_{e2}$ & Excitatory time constant, fast branch & $[3.9,8.4]$ & $\mathrm{ms}$ & HTN \\
$\tau_{i2}$ & Inhibitory time constant, fast branch & $[7.3,16.8]$ & $\mathrm{ms}$ & Conditional uniform \\
$C_1$ & Local coupling factor: pyramidal to excitatory interneurons & $[0.5,1.5]$ & $1$ & HTN \\
$C_2$ & Local coupling factor: excitatory interneurons to pyramidal cells & $[0.4,1.2]$ & $1$ & HTN \\
$C_3$ & Local coupling factor: pyramidal to inhibitory interneurons & $[0.125,0.375]$ & $1$ & HTN \\
$C_4$ & Local coupling factor: inhibitory interneurons to pyramidal cells & $[0.125,0.375]$ & $1$ & HTN \\
$r_{\max}$ & Maximum firing rate & $[2.5,7.5]$ & $\mathrm{s}^{-1}$ & HTN \\
$\theta$ & Sigmoid half-activation potential & $[5.4,6.6]$ & $\mathrm{mV}$ & HTN \\
$\beta$ & Sigmoid slope parameter & $[0.5,0.62]$ & $\mathrm{mV}^{-1}$ & HTN \\
$\delta_s$ & Global delay coordinate (\texttt{delay\_scale}) & $[0,1]$ & $1$ & Uniform \\
\bottomrule
\end{tabular}
\par\smallskip
\begin{minipage}{\linewidth}
\footnotesize
The unit $1$ denotes a dimensionless quantity. HTN denotes the hierarchical truncated normal scheme in Equations~(S1.2)--(S1.3). Inhibitory time constants follow the conditional construction in Equation~(S1.5), within the tabulated bounds. The dimensionless factors $C_1$--$C_4$ are multiplied by $C_{\mathrm{avg}}=135$ inside the simulator. The physical global delay is $d=(10~\mathrm{ms})\delta_s$, with nominal range $[0,10]$~ms. All parameters except $\delta_s$ vary across the 90 regions; $\delta_s$ is shared across the whole brain within each sample.
\end{minipage}
\end{table}

\begin{table}[t]
\centering
\caption{Fixed simulator constants, gain constraints, and structural settings.}
\label{tab:supp_fixed}
\small
\setlength{\tabcolsep}{4pt}
\begin{tabular}{@{}p{0.16\linewidth}p{0.48\linewidth}p{0.17\linewidth}p{0.12\linewidth}@{}}
\toprule
\textbf{Parameter} & \textbf{Description} & \textbf{Value or rule} & \textbf{Unit} \\
\midrule
$H_{e1}/H_{i1}$ & Excitatory/inhibitory synaptic gains, slow branch & $3.25/22$ & $\mathrm{mV}$ \\
$H_{e2}$ & Excitatory synaptic gain, fast branch & $32.5/1000\tau_{e2}$ & $\mathrm{mV}$ \\
$H_{i2}$ & Inhibitory synaptic gain, fast branch & $440/1000\tau_{i2}$ & $\mathrm{mV}$ \\
$C_{\mathrm{avg}}$ & Reference local connectivity scale & 135 & $1$ \\
$\mu/\sigma_p$ & Mean/standard deviation of the Gaussian background input & $220/22$ & $\mathrm{s}^{-1}$ \\
$\sigma_{\mathrm{ref}}$ & Reference amplitude for interregional coupling normalization & $22$ & $\mathrm{s}^{-1}$ \\
$\omega$ & Slow-branch mixing weight (\texttt{band\_w}) & 0.5 & $1$ \\
$\mathbf{K}$ & Structural coupling matrix, fixed across samples & DTI-derived & $1$ \\
$\tau_{\mathrm{delay,base}}$ & Conversion scale for the global delay coordinate & 10 & $\mathrm{ms}$ \\
\bottomrule
\end{tabular}
\par\smallskip
\begin{minipage}{\linewidth}
\footnotesize
The unit $1$ denotes a dimensionless quantity. For paired entries, the listed unit applies to both quantities.Fast-branch gains are determined by the sampled time constants through
$H_{e2}\tau_{e2}=0.0325~\mathrm{mV\,s}$ and
$H_{i2}\tau_{i2}=0.44~\mathrm{mV\,s}$, following the gain rule in
Table~2 of \citet{coroneloliveros2026multifrequency}.
Thus, the products are fixed, whereas the gains vary with the time
constants and are not independently sampled or estimated.
\end{minipage}
\end{table}

\begin{table}[t]
\centering
\caption{Loss weights $w_k$ for multi-task MAE training ($k$ indexes physical targets).}
\label{tab:supp_weights}
\begin{tabular}{l*{11}{c}}
\toprule
Target & $\tau_{e1}$ & $\tau_{i1}$ & $\tau_{e2}$ & $\tau_{i2}$ & $\theta$ & $\beta$ & $r_{\max}$ & $C_1$ & $C_2$ & $C_3$ & $C_4$ \\
\midrule
$w_k$ & 1.2 & 1.0 & 1.5 & 1.5 & 1.0 & 1.5 & 1.0 & 5.0 & 5.0 & 20.0 & 20.0 \\
\bottomrule
\end{tabular}
\end{table}

\begin{table}[t]
\centering
\caption{Hierarchical sampling hyperparameters entering Equations~(S1.2)--(S1.3). For a bound width $\Delta_k=h_k-\ell_k$, the parent-Gaussian standard deviations are $\sigma_{\mathrm{subj}}^{(k)}=\alpha^{\mathrm{subj}}\Delta_k$ and $\sigma_{\mathrm{node}}^{(k)}=\alpha^{\mathrm{node}}\Delta_k$. The choice $\alpha^{\mathrm{subj}}=1/4$ makes the width of the parent Gaussian's $\pm2\sigma$ interval equal to $\Delta_k$. Truncation determines the resulting distribution moments.}
\label{tab:supp_sampler_hp}
\small
\begin{tabular}{@{}p{0.31\linewidth}p{0.10\linewidth}p{0.10\linewidth}p{0.41\linewidth}@{}}
\toprule
\textbf{Parameter family} & $\alpha^{\mathrm{subj}}$ & $\alpha^{\mathrm{node}}$ & \textbf{Notes} \\
\midrule
$\tau_{e1},\;\tau_{e2}$ & $1/4$ & $1/12$ & Smaller within-sample dispersion \\
$\theta,\;\beta,\;r_{\max}$ & $1/4$ & $1/8$ & Intermediate within-sample dispersion \\
$C_1,\;C_2,\;C_3,\;C_4$ & $1/4$ & $1/6$ & Larger within-sample dispersion \\
$\tau_{i1},\;\tau_{i2}$ & --- & --- & Conditional on the matching excitatory time constant; Equation~(S1.4) \\
$\delta_s$ (\texttt{delay\_scale}) & --- & --- & One uniform draw per sample \\
\bottomrule
\end{tabular}
\end{table}

\section*{Extended Methods}

\suppsubsection{Training Data Synthesis: Hierarchical, Physiologically Aware Parameter Sampling}
\label{sec:supp_sampling}

\paragraph{Correspondence with the main-text model}
The state equations and the definitions of the local feedback potential $Z_i(t)$, regional source activity $y_i(t)$, and total afferent input $I_i(t)$ are given in Section~2.2 of the main text. The present section specifies the parameter conventions and sampling procedure used with that model. Local excitatory and inhibitory feedback both depend on $Z_i(t)=\omega v_1^i(t)+(1-\omega)v_4^i(t)$; the inhibitory postsynaptic components $v_3^i$ and $v_6^i$ enter the net source activity through $y_i(t)=\omega[v_2^i(t)-v_3^i(t)]+(1-\omega)[v_5^i(t)-v_6^i(t)]$. The slow and fast branches share the local connectivity factors, sigmoid parameters, and afferent input. The inversion targets comprise 11 region-wise parameter families and one sample-level delay coordinate.

The tabulated $C_1$--$C_4$ are dimensionless factors. The effective local coefficients are $\Gamma_{m,i}=C_{\mathrm{avg}}C_{m,i}$, with $C_{\mathrm{avg}}=135$ and $m=1,\ldots,4$. With $\Phi_i(u)=r_{\max,i}/\{1+\exp[\beta_i(\theta_i-u)]\}$, the main-text feedback functions correspond to
\[
\begin{aligned}
S_{1,i}(u)=S_{4,i}(u)&=\Phi_i(u),\\
S_{2,i}(u)=S_{5,i}(u)&=\Gamma_{2,i}\Phi_i(\Gamma_{1,i}u),\\
S_{3,i}(u)=S_{6,i}(u)&=\Gamma_{4,i}\Phi_i(\Gamma_{3,i}u).
\end{aligned}
\]
The region index on these functions is implicit in the main-text equations. This convention incorporates local connection coefficients in the sigmoid input and output scaling. Membrane potentials and synaptic gains are expressed in mV, firing rates in $\mathrm{s}^{-1}$, and time constants in seconds when evaluating the dynamical equations. Supplementary Table~\ref{tab:supp_sampling} reports time constants in ms.

\paragraph{Fast-branch synaptic gain parameterization}
Following the formulation in \citet{coroneloliveros2026multifrequency}, synaptic gains for the fast branch are parameterized inversely with their corresponding time constants to preserve the area of postsynaptic potentials.

\paragraph{Background input and interregional coupling}
The background term in $I_i(t)$ was generated by independent Gaussian sampling at successive simulation time steps. At $t_n=n\Delta t$, the input was defined as
\[
p_n=\mu+\varepsilon_n,\qquad
\varepsilon_n\overset{\mathrm{i.i.d.}}{\sim}\mathcal{N}(0,\sigma_p^2).
\]
The sampled values were supplied directly to the neural-mass model. Within each simulated sample, a single background sequence was shared across all regions, such that $p_i(t_n)=p_n$. This gives the main-text decomposition $p_i(t_n)=\mu+\widetilde p_i(t_n)$, with $\widetilde p_i(t_n)=\varepsilon_n$. Both local dynamical branches received this background drive through their common afferent input $I_i(t_n)$.

For interregional coupling, $K_{ij}$ denotes the fixed weight from source region $j$ to target region $i$. The effective coefficient $K_{ij}^{*}(t)$ defined in the main text additionally depends on the cumulative history of the source firing rate. Its reference amplitude is fixed at $\sigma_{\mathrm{ref}}=22~\mathrm{s}^{-1}$. The parameter $\sigma_p$ specifies the standard deviation of the Gaussian background input, and $\sigma_{\mathrm{ref}}$ sets the scale of the normalized interregional input. The total input variance depends on the background drive, source fluctuations, and their covariances.

\paragraph{Motivation}
The sampler assigns a dynamical parameter configuration to each synthetic whole-brain sample. The bounds in Supplementary Table~\ref{tab:supp_sampling} define the simulation prior for the reported experiments. The dual-timescale architecture builds on the neural-mass modeling framework described in the main text \citep{david2003modeling}. The fixed gains, sampling intervals, and hierarchy specify the configuration explored in this study. Holding the gains and structural connectivity fixed focuses inversion on local response times, local coupling, sigmoid response properties, and the global transmission timescale. The slow and fast labels refer to the relative synaptic timescales of the two branches; their observable spectra emerge from the coupled nonlinear dynamics.

The dimensionless delay coordinate $\delta_s=\mathrm{delay\_scale}$ is drawn once per sample and shared by all interregional connections:

\begin{equation}
\delta_s \sim \mathrm{Uniform}(0, 1), \quad d_s = \tau_{\mathrm{delay,base}}\delta_s = (10\mathrm{~ms})\delta_s. \tag{S1.1}
\label{eq:supp_delay}
\end{equation}

The nominal physical-delay prior therefore spans $0$--$10$~ms. This scalar is excluded from region-wise FDR analyses. For the regional parameters, a shared sample-level latent location and smaller conditional dispersions induce interregional dependence. Excitatory and inhibitory time constants are additionally coupled through the study-specific constraint in Equation~(S1.4). These choices define a structured prior for synthetic data generation.

\paragraph{Two-Level Truncated Normal Scheme}
Let $s$ index simulated samples, $i$ index brain regions, and $k$ index a directly sampled parameter family. For bounds $[\ell_k,h_k]$ and width $\Delta_k=h_k-\ell_k$, the sampler first draws a sample-level latent location $m_s^{(k)}$ and then draws regional values conditionally independently given that location:

\begin{align}
m_s^{(k)} & \sim \mathcal{TN}\!\left(\frac{\ell_k+h_k}{2}, \, \sigma_{\mathrm{subj}}^{(k)}; \, \ell_k,h_k\right), \tag{S1.2}\label{eq:supp_sample_location}\\
\vartheta_{s,i}^{(k)} \mid m_s^{(k)} &\sim \mathcal{TN}\!\left(m_s^{(k)}, \, \sigma_{\mathrm{node}}^{(k)}; \, \ell_k,h_k\right), \qquad i=1,\ldots,90. \tag{S1.3}\label{eq:supp_regional_draw}
\end{align}

Here $\mathcal{TN}(m,\sigma;\ell,h)$ denotes a normal distribution with parent-Gaussian location $m$ and standard deviation $\sigma$, truncated to $[\ell,h]$. The mean and variance after truncation depend on all four arguments. The hyperparameters are $\sigma_{\mathrm{subj}}^{(k)}=\alpha^{\mathrm{subj}}\Delta_k$ and $\sigma_{\mathrm{node}}^{(k)}=\alpha^{\mathrm{node}}\Delta_k$, with values in Supplementary Table~\ref{tab:supp_sampler_hp}. The label ``subj'' denotes the sample-level component of the synthetic hierarchy.

For $\vartheta=\vartheta_{s,i}^{(k)}$ and $m=m_s^{(k)}$, the law of total variance gives
\begin{equation}
\operatorname{Var}(\vartheta)=
\underbrace{\operatorname{Var}\!\left(\operatorname{E}[\vartheta\mid m]\right)}_{\text{Between-sample component}}
+\underbrace{\operatorname{E}\!\left[\operatorname{Var}(\vartheta\mid m)\right]}_{\text{Within-sample component}}.
\tag{S1.4}\label{eq:supp_total_variance}
\end{equation}
All directly sampled regional families use $\alpha^{\mathrm{subj}}=1/4$. The conditional scale $\alpha^{\mathrm{node}}$ is $1/12$ for excitatory time constants, $1/8$ for sigmoid parameters, and $1/6$ for local connectivity factors. The shared latent location produces exchangeable interregional dependence within each family; anatomical distance is absent from this sampling rule. The global delay is sampled separately according to Equation~\eqref{eq:supp_delay}.).

\paragraph{Coupling Constraints for Inhibitory Time Constants}
For branch $a\in\{1,2\}$, the inhibitory time constant is sampled conditionally on the matching excitatory time constant:
\begin{equation}
\begin{aligned}
\tau_{ia}^{s,i}&=\rho_a^{s,i}\tau_{ea}^{s,i},\\
\rho_a^{s,i}\mid\tau_{ea}^{s,i}
&\sim\mathrm{Uniform}\!\left(
\max\!\left[1,\frac{\ell_{ia}}{\tau_{ea}^{s,i}}\right],\,
\min\!\left[\kappa,\frac{h_{ia}}{\tau_{ea}^{s,i}}\right]\right),
\qquad\kappa=\frac{12}{5}.
\end{aligned}
\tag{S1.5}\label{eq:supp_conditional_tau}
\end{equation}
Here $\ell_{ia}$ and $h_{ia}$ are the branch-specific inhibitory bounds in Supplementary Table~\ref{tab:supp_sampling}. This construction gives the conditional support
\[
\max(\tau_{ea}^{s,i},\ell_{ia})
\leq\tau_{ia}^{s,i}
\leq\min(\kappa\tau_{ea}^{s,i},h_{ia}).
\]
The interval is feasible for the tabulated excitatory bounds. The ordering $\tau_{ia}\geq\tau_{ea}$ and ratio ceiling $\kappa=12/5$ are modeling constraints selected for this study. Under the reported bounds, the maximal achievable ratios $h_{ia} / \ell_{ea}$ (or $h_{ia} / h_{ea}$) confirm that the ratio ceiling $\kappa = 12/5 = 2.4$ exceeds $16.8 / 8.4 = 2.0$ for the fast branch and $35 / 10 = 3.5$ (with $35 / 35 = 1.0$ at the upper bound) for the slow branch, ensuring that the tabulated upper limits remain the active upper restrictions. The inverse-network mapping in Section~S2 uses the same ratio ceiling together with a broader output domain. Following the sampling protocol, we synthesized $S=10^6$ paired parameter--scalp EEG samples for inverse-network training and validation.

\suppsubsection{Training Implementation Details}
\label{sec:supp_training}
Each input sample is a 5~s 19-channel scalp EEG segment sampled at 256~Hz; thus, the network input is represented as
\begin{equation}
    \mathbf{X}\in\mathbb{R}^{B\times C\times T},\qquad C=19,\quad T=5\times256=1280,
    \tag{S2.1}
\end{equation}
where $B$ denotes the mini-batch size. Each EEG segment undergoes a two-step demeaning procedure: first subtracting the temporal mean per channel, and then subtracting the cross-channel spatial mean at each time point, enforcing an approximate zero mean along both axes. For numerical stability, segments are screened by peak-to-peak amplitude within the analysis window, retaining only traces within $[3,\,100]~\mu$V.

The inverse mapper of NeuroDyn-EEG employs a multi-branch spatial--spectral inversion (MB-SSI) architecture. This architecture preserves three complementary streams of feature information in parallel: the scalp temporal branch captures multi-scale waveforms and long-range temporal contexts; the source-space branch constructs localized representations indexed by AAL nodes via the leadfield pseudoinverse; and the spectral branch explicitly encodes low-frequency oscillatory profiles. Subsequently, taking source node tokens as queries and scalp temporal tokens as keys and values, the model integrates inter-nodal and cross-domain information through node self-attention and cross-attention. Attention weights are not interpreted as anatomical connectivity, effective connectivity, or physical electrode sensitivities. The 12-target configuration used in the current study contains approximately $2.43\times10^6$ trainable parameters. Layer-by-layer specifications and tensor dimensions are detailed in Supplementary Table~\ref{tab:supp_neurodyn_architecture}.

{\small
\setlength{\tabcolsep}{3pt}
\renewcommand{\arraystretch}{1.08}
\begin{longtable}{p{0.16\textwidth}p{0.30\textwidth}p{0.25\textwidth}p{0.16\textwidth}}
\caption{Layer-wise specification of the NeuroDyn-EEG multi-branch spatial--spectral inversion network. $B$ denotes the batch size. Tensor dimensions follow the channel-first convention.}
\label{tab:supp_neurodyn_architecture}\\
\toprule
\textbf{Module} & \textbf{Layer / operation} & \textbf{Parameters} & \textbf{Output size} \\
\midrule
\endfirsthead
\multicolumn{4}{l}{\small Table~\ref{tab:supp_neurodyn_architecture} continued from the previous page.}\\
\toprule
\textbf{Module} & \textbf{Layer / operation} & \textbf{Parameters} & \textbf{Output size} \\
\midrule
\endhead
\bottomrule
\endfoot

\multirow{3}{0.16\textwidth}{\textbf{Input and preprocessing}}
& Scalp EEG segment & $C=19$, $f_s=256$~Hz, $T=1280$ (5~s) & $B\times19\times1280$ \\
\cmidrule(l){2-4}
& Temporal and spatial demeaning & Channel-wise temporal mean removal; sample-wise cross-channel mean removal & $B\times19\times1280$ \\
\cmidrule(l){2-4}
& Peak-to-peak amplitude screening & Retained range: $[3,100]~\mu$V & $B\times19\times1280$ \\
\midrule

\multirow{7}{0.16\textwidth}{\textbf{Scalp temporal branch}}
& Four parallel Conv1D paths & $C_{\mathrm{in}}=19$, $C_{\mathrm{out}}=32$ per path; $k\in\{3,5,7,9\}$; $s=2$; same padding; no bias & $4\times(B\times32\times640)$ \\
\cmidrule(l){2-4}
& Channel concatenation & $4\times32=128$ channels & $B\times128\times640$ \\
\cmidrule(l){2-4}
& Dilated residual block 1 & Two Conv1D layers; $C=128$, $k=5$, $d=2$; pre-BN; LeakyReLU$(0.2)$; dropout $=0.1$ & $B\times128\times640$ \\
\cmidrule(l){2-4}
& Dilated residual block 2 & Two Conv1D layers; $C=128$, $k=5$, $d=4$; pre-BN; LeakyReLU$(0.2)$; dropout $=0.1$ & $B\times128\times640$ \\
\cmidrule(l){2-4}
& Adaptive average pooling & $640\rightarrow8$ temporal bins & $B\times128\times8$ \\
\cmidrule(l){2-4}
& Pointwise projection & Conv1D, $k=1$, $128\rightarrow192$ & $B\times192\times8$ \\
\cmidrule(l){2-4}
& Transposition and sinusoidal positional encoding & $d_{\mathrm{model}}=192$, sequence length $=8$ & $B\times8\times192$ \\
\midrule

\multirow{6}{0.16\textwidth}{\textbf{Source pseudo-inverse branch}}
& Tikhonov-regularized inverse projection & $\mathbf{L}^{+}\in\mathbb{R}^{90\times19}$; ridge $\lambda=10^{-3}$ & $B\times90\times1280$ \\
\cmidrule(l){2-4}
& Grouped Conv1D + BN + LeakyReLU & $90\rightarrow1080$; $k=7$, $s=4$, $p=3$, groups $=90$ & $B\times1080\times320$ \\
\cmidrule(l){2-4}
& Grouped Conv1D + BN + LeakyReLU & $1080\rightarrow1080$; $k=5$, $s=4$, $p=2$, groups $=90$ & $B\times1080\times80$ \\
\cmidrule(l){2-4}
& Grouped Conv1D + BN + LeakyReLU & $1080\rightarrow2160$; $k=3$, $s=2$, $p=1$, groups $=90$ & $B\times2160\times40$ \\
\cmidrule(l){2-4}
& Adaptive average pooling & $40\rightarrow1$ temporal bin & $B\times2160\times1$ \\
\cmidrule(l){2-4}
& Reshape to node-wise features & 24 features per AAL node & $B\times90\times24$ \\
\midrule

\multirow{7}{0.16\textwidth}{\textbf{Spectral branch}}
& Real-valued FFT & 1280 time samples; 641 non-negative-frequency bins & $B\times19\times641$ \\
\cmidrule(l){2-4}
& Log-magnitude transform and frequency truncation & $\log(1+|\operatorname{rFFT}(\mathbf{X})|)$; first 201 bins ($0$--$40$~Hz) & $B\times19\times201$ \\
\cmidrule(l){2-4}
& Conv1D + BN + LeakyReLU & $19\rightarrow32$; $k=5$, $s=1$, $p=2$ & $B\times32\times201$ \\
\cmidrule(l){2-4}
& Conv1D + BN + LeakyReLU & $32\rightarrow32$; $k=5$, $s=2$, $p=2$ & $B\times32\times101$ \\
\cmidrule(l){2-4}
& Conv1D + BN + LeakyReLU & $32\rightarrow64$; $k=3$, $s=2$, $p=1$ & $B\times64\times51$ \\
\cmidrule(l){2-4}
& Adaptive average pooling and flattening & $51\rightarrow1$ frequency bin & $B\times64$ \\
\cmidrule(l){2-4}
& Spectral projection & Linear, $64\rightarrow64$ & $B\times64$ \\
\midrule

\multirow{5}{0.16\textwidth}{\textbf{Node-token construction}}
& Learnable node embedding & 90 embeddings, $d_{\mathrm{node}}=32$ & $B\times90\times32$ \\
\cmidrule(l){2-4}
& Feature concatenation & $24+32=56$ features per node & $B\times90\times56$ \\
\cmidrule(l){2-4}
& Node projection & Linear, $56\rightarrow192$ & $B\times90\times192$ \\
\cmidrule(l){2-4}
& Learnable global query expansion & One query, $d_{\mathrm{model}}=192$ & $B\times1\times192$ \\
\cmidrule(l){2-4}
& Token concatenation & 90 node tokens + 1 global token & $B\times91\times192$ \\
\midrule

\multirow{2}{0.16\textwidth}{\textbf{Dense node-token self-attention} $\boldsymbol{\times2}$}
& Pre-LN multi-head self-attention + residual & 6 heads; head dimension $=32$; dropout $=0.1$ & $B\times91\times192$ \\
\cmidrule(l){2-4}
& Pre-LN feed-forward network + residual & $192\rightarrow384\rightarrow192$; GELU; dropout $=0.1$ & $B\times91\times192$ \\
\midrule

\multirow{2}{0.16\textwidth}{\textbf{Scalp-to-source cross-attention} $\boldsymbol{\times2}$}
& Pre-LN multi-head cross-attention + residual & $Q:91$ source/global tokens; $K,V:8$ scalp-time tokens; 6 heads; head dimension $=32$ & $B\times91\times192$ \\
\cmidrule(l){2-4}
& Pre-LN feed-forward network + residual & $192\rightarrow384\rightarrow192$; GELU; dropout $=0.1$ & $B\times91\times192$ \\
\midrule

\multirow{10}{0.16\textwidth}{\textbf{Parameter-specific prediction heads}}
& Broadcast spectral feature over nodes & 64 spectral features per node & $B\times90\times64$ \\
\cmidrule(l){2-4}
& Node--spectral feature concatenation & $192+64=256$ features per node & $B\times90\times256$ \\
\cmidrule(l){2-4}
& Node-head hidden layer 1 & Independent per parameter; Linear $256\rightarrow128$; LeakyReLU$(0.2)$; dropout $=0.1$ & $B\times90\times128$ \\
\cmidrule(l){2-4}
& Node-head hidden layer 2 & Linear $128\rightarrow128$; LeakyReLU$(0.2)$ & $B\times90\times128$ \\
\cmidrule(l){2-4}
& Node-head output layer & Linear $128\rightarrow1$; squeeze last dimension & $B\times90$ per target \\
\cmidrule(l){2-4}
& Global--spectral feature concatenation & $192+64=256$ features & $B\times256$ \\
\cmidrule(l){2-4}
& Global-head hidden layer & Linear $256\rightarrow128$; LeakyReLU$(0.2)$; dropout $=0.1$ & $B\times128$ \\
\cmidrule(l){2-4}
& Global-head output layer & Linear $128\rightarrow1$ & $B\times1$ \\
\cmidrule(l){2-4}
& Physiological output constraints & Soft-bounded sigmoid or $\tau_i$--$\tau_e$ coupled mapping; $\kappa=12/5$ & $B\times90$ or $B\times1$ \\
\cmidrule(l){2-4}
& Output dictionary & 11 node-wise targets + 1 sample-wise target & $11\times(B\times90)$ and $B\times1$ \\
\end{longtable}
}

In the source-space branch, the regularized pseudoinverse of the leadfield matrix $\mathbf{L}\in\mathbb{R}^{19\times90}$ is fixed and is not trainable, but is registered as a fixed buffer saved with the model:
\begin{equation}
    \mathbf{L}^{+}=\left(\mathbf{L}^{\mathsf T}\mathbf{L}+\lambda\mathbf{I}\right)^{-1}\mathbf{L}^{\mathsf T},
    \qquad \lambda=10^{-3}.
    \tag{S2.2}
\end{equation}
The resulting 90 source signals are individually processed by grouped 1D convolutions with the number of groups strictly set to 90, ensuring that no convolutional kernel mixes information across distinct brain regions at this stage. Three convolutional layers sequentially compress the temporal length from $1280$ to $320$, $80$, and $40$, followed by temporal average pooling to yield a 24-dimensional feature vector per node. After concatenation with a 32-dimensional learnable positional embedding per node, a linear layer projects the 56-dimensional input to $d_{\mathrm{model}}=192$. This design concurrently incorporates physical localization dictated by the leadfield and data-driven node identity encoding.

The scalp temporal branch jointly encodes the 19 channels into 128 feature channels. Four parallel convolutions with kernel sizes of 3, 5, 7, and 9 initially halve the temporal length; subsequently, two residual blocks with dilation rates of 2 and 4 refine the representations. Adaptive average pooling compresses the full 5~s window into eight ordered temporal slots, which, after $1\times1$ convolutional projection and addition of fixed sinusoidal positional encodings, form the scalp temporal memory $\mathbf{K}=\mathbf{V}\in\mathbb{R}^{B\times8\times192}$.

The spectral branch computes a real FFT along the temporal axis of the same input and applies a $\log(1+|\cdot|)$ compression to the magnitudes. The network retains the first 201 non-negative-frequency bins, including the DC component ($k=0,\ldots,200$). For 1280 samples acquired at 256~Hz, the frequency spacing is $\Delta f=256/1280=0.2$~Hz, and the retained bin centers span $0$--$40$~Hz. Three convolutional layers, global average pooling, and linear projection produce a 64-dimensional sample-level spectral vector. This vector is broadcast across all brain regions at the prediction stage and concatenated with each final node token, providing explicit low-frequency oscillatory conditioning for dynamical parameter inversion.

The 90 source node tokens and one learnable global query token first pass through two self-attention layers. Each layer employs pre-layer normalization, 6-head attention, residual connections, and a feed-forward network with an intermediate dimension of 384; per-head dimension is $192/6=32$. Because no fixed adjacency mask is enforced, the 90 nodes constitute a fully connected graph, where the attention matrix functions as a sample-adaptive dense relational operator. Subsequently, two cross-attention layers of identical architecture take these 91 tokens as queries and the eight scalp temporal tokens as keys and values:
\begin{gather}
    \mathbf{A} = \operatorname{softmax}\!\left(\frac{(\mathbf{Q}\mathbf{W}_{Q})(\mathbf{K}\mathbf{W}_{K})^{\mathsf T}}{\sqrt{32}}\right), \notag\\
    \operatorname{CrossAttn}(\mathbf{Q},\mathbf{K},\mathbf{V}) = \mathbf{A}(\mathbf{V}\mathbf{W}_{V}),
    \tag{S2.3}
\end{gather}
with residual connections applied after each attention and feed-forward sublayer. This asymmetric fusion enables source nodes to actively select the most relevant scalp temporal evidence for their parameter estimation, while the global token aggregates sample-level information for predicting $\mathrm{delay\_scale}$.

The fused node tokens are concatenated with the broadcast spectral vector to form 256-dimensional features. The 11 region-wise parameter families each utilize an independent 3-layer MLP to generate $B\times90$ raw outputs; the global delay scaling factor $\mathrm{delay\_scale}$ utilizes a shallower 2-layer MLP to generate a $B\times1$ output from the global token. The model does not share final parameter heads, allowing distinct physical quantities to learn tailored inverse mappings. For all parameters other than the inhibitory time constants, the raw output $a_k$ is mapped via a soft-bounded Sigmoid:
\begin{equation}
    \widehat{\theta}_k=\ell_k+(h_k-\ell_k)
    \left[(1+2\alpha_k)\sigma(a_k)-\alpha_k\right],
    \tag{S2.4}
\end{equation}
where $[\ell_k,h_k]$ is the physiological prior interval, and $\alpha_k\geq0$ controls the allowable extension relative to the interval width; $\alpha_k=0$ recovers the strictly bounded Sigmoid. For the two sets of inhibitory time constants, the network employs a dedicated output layer coupled with the matching excitatory time constant:
\begin{equation}
    \widehat{\tau}_{i\bullet}=\ell_{i\bullet}+
    \left(\frac{12}{5}\widehat{\tau}_{e\bullet}-\ell_{i\bullet}\right)
    \sigma(a_{i\bullet}),\qquad \bullet\in\{1,2\},
    \tag{S2.5}
\end{equation}
This mapping incorporates the study-specific ratio ceiling $\kappa=12/5$ into the output parameterization. As specified in Section~S1, this ratio ceiling is a modeling choice; the conditional training prior additionally enforces absolute time-constant bounds and the ordering $\tau_i\geq\tau_e$. Target bounds and multi-task loss weights are provided in Supplementary Table~\ref{tab:supp_sampling} and Supplementary Table~\ref{tab:supp_weights}, respectively.

The model is trained using the Adam optimizer with an initial learning rate of $1 \times 10^{-3}$, a weight decay of $1 \times 10^{-5}$, a batch size of 256, and a maximum of 150 training epochs. The generated corpus is split $5{:}1$ into training and validation sets. To prevent overfitting, an early stopping mechanism terminates training if the validation loss does not improve for 20 consecutive epochs, and the checkpoint with the lowest validation loss is selected. The model is implemented in Python 3.7.12 using PyTorch 1.7.1, with all experiments conducted on a server equipped with 4 NVIDIA T4 GPUs (16 GB VRAM each) accelerated by CUDA 11.0. The complete training and validation pipeline is summarized in Supplementary Algorithm~\ref{alg:supp_train}.

\begin{algorithm}[H]
\caption{Training and Validation Procedure for EEG-to-Parameter Inversion}
\label{alg:supp_train}
\begin{algorithmic}[1]
\Require Paired training data $\mathcal{D}_{\mathrm{train}}$ and validation data $\mathcal{D}_{\mathrm{val}}$; parameter set $\mathcal{P}$; loss weights $\{w_k\}_{k\in\mathcal{P}}$
\Require Batch size $B=256$; maximum epochs $E=150$; learning rate $\eta=10^{-3}$; weight decay $\lambda_{\mathrm{wd}}=10^{-5}$; early stopping patience $P=20$
\State Initialize model $f_\psi$; initialize Adam optimizer with learning rate $\eta$ and weight decay $\lambda_{\mathrm{wd}}$
\State $\mathcal{L}_{\mathrm{best}} \leftarrow +\infty$; $c_{\mathrm{noimp}} \leftarrow 0$
\For{$\mathit{epoch} \leftarrow 1$ \textbf{to} $E$}
    \For{each mini-batch $(x, \theta)$ of size $B$ sampled from $\mathcal{D}_{\mathrm{train}}$}
        \State Demean $x$ along both temporal and spatial axes
        \State Select valid index set $I = \{i \mid 3 \leq \mathrm{amp}(x_i) \leq 100\}$
        \If{$|I| = 0$} \State \textbf{continue to next mini-batch} \EndIf
        \State $x \leftarrow x[I]$, $\theta \leftarrow \theta[I]$
        \State $\mathcal{L} \leftarrow \displaystyle\sum_{k\in\mathcal{P}} w_k \cdot \mathrm{MAE}(f_\psi(x)_k,\,\theta_k)$
        \State Update $\psi$ via Adam according to $\nabla_\psi \mathcal{L}$
    \EndFor
    \State Compute weighted validation loss $\mathcal{L}_{\mathrm{val}}$ on $\mathcal{D}_{\mathrm{val}}$
    \If{$\mathcal{L}_{\mathrm{val}} < \mathcal{L}_{\mathrm{best}}$}
        \State Save checkpoint $f_\psi$; $\mathcal{L}_{\mathrm{best}} \leftarrow \mathcal{L}_{\mathrm{val}}$; $c_{\mathrm{noimp}} \leftarrow 0$
    \Else
        \State $c_{\mathrm{noimp}} \leftarrow c_{\mathrm{noimp}} + 1$
        \If{$c_{\mathrm{noimp}} \geq P$}
            \State \textbf{Terminate training early}
            \State \textbf{break}
        \EndIf
    \EndIf
\EndFor
\State \Return Checkpoint $f_\psi$ with lowest validation loss
\end{algorithmic}
\end{algorithm}

\suppsubsection{Downstream Preprocessing: Baseline and NeuroDyn-EEG Cohort Rules}
\label{sec:supp_preprocess}
\paragraph{Baselines}
Downstream evaluations strictly adhere to each baseline model's officially published preprocessing pipeline, ensuring that performance variations are attributable to architecture rather than customized uniform filtering:

\begin{description}
    \item[\textbf{CBraMod}:] 19 standard 10--20 channels; 0.3--75 Hz bandpass and 60 Hz notch filters; resampled to 200 Hz; 30-second non-overlapping segments; segments with amplitude $>$ $\pm100~\mu$V discarded; amplitudes scaled to $[-1,1]$ by a factor of 100.
    \item[\textbf{LaBraM}:] 0.1--75 Hz bandpass, 50 Hz notch; resampled to 200 Hz; inputs scaled to $[-1,1]$ in units of 0.1 mV.
    \item[\textbf{BrainOmni}:] 0.1--96 Hz bandpass, 50/60 Hz notch; resampled to 256 Hz; bad channels detected via PSD and interpolated; common average reference (CAR); channel-wise z-scoring.
    \item[\textbf{VAEEG}:] Resampled to 256 Hz; segments $>400~\mu$V rejected; CAR; FFT band power across five frequency bands (Delta to High Beta); 1-second segments fed into the frequency-band autoencoder.
\end{description}

\paragraph{NeuroDyn-EEG (Clinical Harmonization)}
The inverse network expects 19 standard 10--20 electrode channels and a sampling grid consistent with the forward model and synthetic pre-training corpus in the main text. For each public cohort, we perform the following steps: (i)~select the standard sensor layout; (ii)~apply band-limiting and resampling according to acquisition constraints; (iii)~apply channel-wise temporal mean removal, sample-wise cross-channel spatial mean removal, and peak-to-peak amplitude screening within the analysis window (matching the quality control categories in Section~S2); (iv)~register retained segments per subject, ensuring that classification evaluation and JR parameter group comparisons are always conducted on identical screened data and clinical labels.

\suppsubsection{Downstream Evaluation Setup}
\label{sec:supp_eval}
We employ subject-stratified 5-fold cross-validation: Within each fold, the training, validation, and test subject sets are mutually exclusive..
\begin{itemize}
    \item \textbf{Custom Splits (AD65, PD31, MDD):} Subjects are partitioned into five disjoint subsets; each fold uses three subsets for training, one for validation, and one for testing.
    \item \textbf{TUAB (Official Split):} The official training/evaluation split is preserved; the official training set is further divided into 5 folds (4 training / 1 validation per fold); the official evaluation set serves as the final test set.
\end{itemize}
To account for class imbalance, we evaluate models using Balanced Accuracy (BACC), the Area Under the Receiver Operating Characteristic Curve (AUCROC), and the Area Under the Precision-Recall Curve (AUCPR). Notably, BACC is computed as macro-averaged recall, while AUCPR places greater emphasis on performance for the minority positive class. Two independent random seeds $\times$ 5 folds generate 10 runs per task; checkpoints are selected based on peak validation BACC; test metrics are reported as mean $\pm$ standard deviation across runs.

\suppsubsection{Simulation Experiment Metrics}
\label{sec:supp_simmetrics}

\textbf{Parameter Recovery (Experiment~1)} utilizes the \textbf{Pearson correlation coefficient} computed on pooled predicted vs.\ true values for each model output target, as well as the \textbf{Pearson correlation coefficient} across each cortical region for spatially varying quantities shown in the main text heatmap subplots. The 11 JR parameter families are defined across AAL-90 brain regions; $\mathrm{delay\_scale}$ is a sample-level scalar.

\noindent\textbf{Surrogate Noise Sensitivity (Experiment~2)} For each target parameter $p$, surrogate noise family $k$, and SNR level $s$, Pearson $r$, coefficient of determination $R^2$, and mean absolute error (MAE) are computed across $N$ evaluation samples. All three metrics are calculated independently at the parameter level without pooling the 12 targets prior to evaluation. Pearson $r$ evaluates the linear consistency between predictions and ground truths; $R^2$ and MAE are defined as
\begin{equation*}
\begin{aligned}
\mathrm{SSE}_{p,k,s}
&= \sum_{i=1}^{N}\left(\theta_{i,p}-\hat{\theta}_{i,p}^{(k,s)}\right)^2, \\
\mathrm{SST}_{p}
&= \sum_{i=1}^{N}\left(\theta_{i,p}-\bar{\theta}_{p}\right)^2, \\
R^2_{p,k,s}
&= 1-\frac{\mathrm{SSE}_{p,k,s}}{\mathrm{SST}_{p}}, \\
\mathrm{MAE}_{p,k,s}
&= \frac{1}{N}\sum_{i=1}^{N}\left|\theta_{i,p}-\hat{\theta}_{i,p}^{(k,s)}\right|,
\end{aligned}
\end{equation*}
where $\bar{\theta}_{p}$ denotes the mean ground-truth value of parameter $p$ across evaluation samples. $R^2=1$ indicates perfect prediction; $R^2=0$ indicates that the squared prediction error equals the baseline error of predicting the ground-truth mean; when $\mathrm{SSE}>\mathrm{SST}$, $R^2<0$. A lower MAE indicates smaller absolute prediction error. Because distinct parameters possess different physical dimensions and numerical scales, raw MAE is solely used within the same parameter to compare noise types, SNR levels, and clean baselines, rather than for comparing recovery difficulty across different parameters. Parameter-wise $R^2$ and MAE results are presented in Supplementary Figure~\ref{fig:supp_noise_r2} and Supplementary Figure~\ref{fig:supp_noise_mae}, respectively; lower SNR corresponds to stronger contamination. These phenomenological surrogate noise models are designed for stress-testing and do not encompass the full complexity of real-world artifacts.

\begin{figure}[p]
    \centering
    \includegraphics[width=\textwidth]{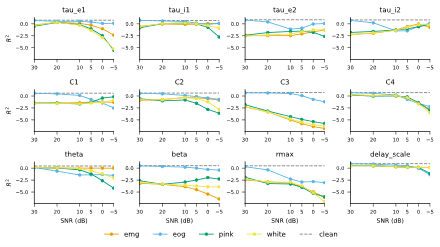}
    \caption{\textbf{Parameter-wise $R^2$ robustness results under four surrogate noise types.}
    The 12 subpanels correspond to the 12 model output targets; each subpanel presents $R^2$ values for white noise, pink noise, EMG-like noise, and EOG-like noise across SNR 30 to $-5$\,dB, referencing the clean baseline result for that parameter. $R^2=1$ indicates perfect agreement; $R^2=0$ indicates that prediction mean squared error equals the ground-truth mean baseline error; $R^2<0$ indicates residual sum of squares exceeds that baseline error. All curves are computed independently per parameter.}
    \label{fig:supp_noise_r2}
\end{figure}

\begin{figure}[p]
    \centering
    \includegraphics[width=\textwidth]{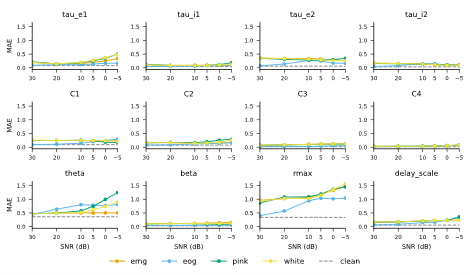}
    \caption{\textbf{Parameter-wise MAE robustness results under four surrogate noise types.}
    Subpanels, noise types, SNR levels, and clean baseline references are identical to Supplementary Figure~\ref{fig:supp_noise_r2}. Lower MAE indicates smaller absolute prediction error; the vertical axis preserves the original physical scale of each parameter. Therefore, curves should be used to compare different noise types and SNR conditions within each parameter panel, rather than comparing recovery difficulty across distinct parameters based on absolute MAE magnitude.}
    \label{fig:supp_noise_mae}
\end{figure}

\noindent\textbf{Closed-Loop Evaluation (Experiment~3)} evaluates inverse--forward closed-loop reconstruction on real resting-state EEG: observed waveforms are inverted and injected back into the forward JR simulator to yield reconstructed waveforms, followed by consistency evaluation. Temporal consistency is assessed via Phase-Locking Value (PLV) across $\delta$ (1--4\,Hz), $\theta$ (4--8\,Hz), and $\alpha$ (8--13\,Hz) bands; the $\beta$ band is excluded from PLV analysis due to susceptibility to aperiodic $1/f$ backgrounds and scalp EMG, though its power profile remains covered by spectral metrics. Spectral consistency within 1--40\,Hz is quantified via three metrics: Pearson correlation of log Welch power spectra ($r_{\log}$), absolute $\alpha$-peak frequency error within 7--13\,Hz, and absolute error in aperiodic $1/f$ spectral slope in log-log space; channel-wise metrics are averaged across subjects.

\suppsubsection{Noise Synthesis and SNR Calibration}
\label{sec:supp_noise}

\paragraph{Phenomenological Noise Modeling and Robustness Benchmarking}
Real scalp EEG is subject to diverse physiological and non-physiological interferences, contributing to distribution shifts between empirical recordings and clean simulations. To quantify the model's sensitivity to canonical out-of-distribution perturbations, we construct an SNR-calibrated semi-synthetic noise injection framework. This framework selects four phenomenological proxies: (i) Gaussian white noise approximating sensor and amplifier thermal noise floors; (ii) aperiodic, scale-free ``$1/f$'' components \citep{donoghue2020parameterizing}; (iii) electromyographic (EMG)-like artifacts with energy concentrated in higher frequency bands \citep{goncharova2003spectral,muthukumaraswamy2013high}; and (iv) electrooculographic (EOG)-like artifacts featuring low-frequency baseline drifts and high-amplitude transients \citep{croft2000eog}. These proxies do not span the entire spectrum of empirical artifacts.

In the experimental protocol, additive noise is injected into clean forward-simulated signals $\mathbf{x} \in \mathbb{R}^{C \times T}$ ($C=19$ channels, sampling rate $f_s=256\,\mathrm{Hz}$) prior to parameter inversion. This study adopts \textbf{phenomenological modeling} to approximate canonical time-frequency characteristics directly in sensor space. The ocular component draws on the established biophysical characterization of the EOG \citep{elbert1985removal}. This additive noise protocol at specified SNR levels parallels stress-testing paradigms in EEG deep denoising benchmarks \citep{zhang2021eegdenoisenet}, offering transparent implementation and reproducibility while serving as a phenomenological proxy rather than a high-fidelity biophysical simulation.

\paragraph{Signal-to-Noise Ratio Calibration (Per-Trial)}
For each 5-second segment, \textbf{signal power} is defined as the mean squared amplitude across channels and time samples:
\begin{equation}
P_{\mathrm{sig}} = \frac{1}{CT} \sum_{c=1}^C \sum_{t=1}^T x_{c,t}^2. \tag{S6.1}\label{eq:supp_psig}
\end{equation}
Given a target \(\mathrm{SNR}_{\mathrm{dB}}\), the required \textbf{noise power} follows the standard power ratio:
\begin{equation}
P_{\mathrm{noise}} = \frac{P_{\mathrm{sig}}}{10^{\mathrm{SNR}_{\mathrm{dB}} / 10}}, \tag{S6.2}\label{eq:supp_pnoise}
\end{equation}
or equivalently \(\mathrm{SNR}_{\mathrm{dB}} = 10\log_{10}(P_{\mathrm{sig}}/P_{\mathrm{noise}})\). For each surrogate family, an unscaled template \(\tilde{\boldsymbol{\epsilon}}\) is generated, and its mean squared amplitude is computed:
\begin{equation}
P_{\mathrm{raw}} = \frac{1}{CT} \sum_{c,t} \tilde{\epsilon}_{c,t}^2. \tag{S6.3}\label{eq:supp_praw}
\end{equation}
The calibrated noise is then scaled as:
\begin{equation}
\boldsymbol{\epsilon} = \sqrt{\frac{P_{\mathrm{noise}}}{P_{\mathrm{raw}}}} \tilde{\boldsymbol{\epsilon}}. \tag{S6.4}\label{eq:supp_noise_scale}
\end{equation}

The contaminated segment is given by \(\mathbf{x}_{\mathrm{noisy}} = \mathbf{x} + \boldsymbol{\epsilon}\). Calibration is performed on a \textbf{per-trial} basis to ensure comparability across segments with heterogeneous RMS amplitudes at fixed \(\mathrm{SNR}_{\mathrm{dB}}\). Evaluated SNR levels are \(\mathrm{SNR}_{\mathrm{dB}}\in\{30,20,10,5,0,-5\}\), with lower \(\mathrm{SNR}_{\mathrm{dB}}\) corresponding to stronger contamination. For each noise type and \(\mathrm{SNR}_{\mathrm{dB}}\), \(\mathbf{x}_{\mathrm{noisy}}\) is processed by the inverse network, and \(\hat{\boldsymbol{\theta}}\) is evaluated against ground-truth parameters.

\paragraph{Gaussian White Noise}
Gaussian white noise models instrumental thermal noise and basic background noise floors at the electrode--scalp interface \citep{uriguen2015eeg}. Statistically, it is modeled as a discrete-time stochastic process whose samples at any time $t$ are independent and identically distributed according to $x_{\mathrm{white}}(t) \sim \mathcal{N}(0, \sigma_{\mathrm{white}}^2)$. Due to temporal memorylessness, white noise exhibits a uniform power spectral density across all frequencies. In algorithm simulations, modulating the variance $\sigma_{\mathrm{white}}^2$ provides precise control over the SNR level of background instrumentation noise, assessing parameter estimation stability under varying degrees of baseline interference.

\paragraph{Pink ($1/f$-Type) Noise}
This component approximates the aperiodic, scale-free background activity observed in empirical resting-state EEG \citep{he2014scale}. Unlike coherent neural oscillations with distinct peak frequencies, the power spectral density $S_{\mathrm{pink}}(f)$ of pink noise exhibits a power-law decay with frequency $f$ \citep{donoghue2020parameterizing}, i.e., $S_{\mathrm{pink}}(f) \propto 1/f^\alpha$ (set to $\alpha \approx 1$). Implementation-wise, we employ frequency-domain filtering: generating a standard Gaussian white noise sequence $x_{\mathrm{white}}(t)$, mapping it to the frequency domain via discrete Fourier transform ($\mathcal{F}$), applying an amplitude decay filter $f^{-1/2}$, and transforming back to the time domain via inverse Fourier transform ($\mathcal{F}^{-1}$):
\begin{equation}
x_{\mathrm{pink}}(t) = \mathcal{F}^{-1}\{\mathcal{F}\{x_{\mathrm{white}}(t)\} \cdot f^{-1/2}\}. \tag{S6.5}\label{eq:supp_pink}
\end{equation}
This component tests model sensitivity to low-frequency, high-energy background perturbations.

\paragraph{EMG-Like Noise}
Scalp electromyographic (EMG) activity arises from facial, cranial, and cervical muscle contractions, characterized by temporal burstiness and high-frequency spectral energy \citep{goncharova2003spectral}. To approximate these burst dynamics, we model EMG-like artifacts as envelope-modulated bandpass stochastic processes \citep{muthukumaraswamy2013high}:
\begin{equation}
x_{\mathrm{EMG}}(t) = A(t) \cdot [h_{\mathrm{bp}}(t) * w(t)], \tag{S6.6}\label{eq:supp_emg}
\end{equation}
where $w(t) \sim \mathcal{N}(0, 1)$ represents broadband Gaussian driving noise; $h_{\mathrm{bp}}(t)$ denotes the impulse response of a bandpass filter (passband 20--100\,Hz); and $*$ denotes convolution. The envelope $A(t)$ is a slowly varying non-negative modulation signal generated by low-pass smoothing a sparse Poisson-like random pulse train, capturing temporal stochasticity and burst energy dynamics associated with muscle activation.

\paragraph{EOG-Like Noise}
Guided by the characteristic features of ocular contamination described in the EEG artifact literature \citep{elbert1985removal,croft2000eog}, we formulate a simplified phenomenological model as a linear superposition of two ocular components:
\begin{enumerate}
    \item \textbf{Slow Drift}: Capturing high-amplitude, slow ocular movements volume-conducted to the scalp, obtained by applying a low-pass filter with an absolute cutoff of $4\,\mathrm{Hz}$ to standard Gaussian white noise \citep{croft2000eog}:
    \begin{equation}
    x_{\mathrm{drift}}(t) = \mathcal{F}^{-1}\{\mathcal{F}\{x_{\mathrm{white}}(t)\} \cdot H_{\mathrm{LP}\le 4\mathrm{Hz}}(f)\}. \tag{S6.7}\label{eq:supp_eog_drift}
    \end{equation}
    \item \textbf{Transient Bursts}: Modeling eyeblink artifacts. Random onset times $t_{0,i}$ are generated at an average rate of $\approx 0.2\,\mathrm{Hz}$ (approximately once every 5 seconds), modeled as Hann-window pulses with durations $T_i \in [0.1, 0.4]\,\mathrm{s}$:
    \begin{equation}
    x_{\mathrm{burst}}(t) = \sum_i A_i \cdot \frac{1}{2} \left[1 - \cos\left(\frac{2\pi(t - t_{0,i})}{T_i}\right)\right] \cdot \mathbb{I}_{[t_{0,i}, t_{0,i}+T_i]}(t), \tag{S6.8}\label{eq:supp_eog_burst}
    \end{equation}
    where $A_i$ is a random amplitude scaling factor and $\mathbb{I}$ is the indicator function. This formulation matches the characteristic morphology targeted by classical regression and template-matching blink removal algorithms \citep{croft2000eog,li2006blink}.
\end{enumerate}
Before global SNR scaling, the composite EOG template is formed as $x_{\mathrm{EOG}}(t) = x_{\mathrm{drift}}(t) + 0.5 \times x_{\mathrm{burst}}(t)$.

\suppsubsection{Clinical Cohort JR Parameter Statistics and AAL Naming Mapping}
\label{sec:supp_jr_stats}
\paragraph{AD65}
Under \(q_{\mathrm{BH}}<0.001\), the parameter family exhibiting the largest number of significant brain regions in AD65 is \texttt{$C_1$} (37/90 AAL regions), followed by \texttt{$\tau_{e2}$} (33), \texttt{$\theta$} (22), \texttt{$\tau_{e1}$} (15), and \texttt{$\beta$} (14). According to the AAL index mapping, the 37 significant regions for \texttt{$C_1$} are:
R Precentral Gyrus, L Superior Frontal Gyrus, L Superior Orbital Gyrus, R Superior Orbital Gyrus, L Middle Orbital Gyrus, L IFG (p. Opercularis), R IFG (p. Opercularis), L IFG (p. Triangularis), L IFG (p. Orbitalis), L Posterior-Medial Frontal, L Olfactory cortex, R Olfactory cortex, L Insula Lobe, L Cingulum Ant, L Cingulum Mid, L Hippocampus, R Hippocampus, L ParaHippocampal Gyrus, L Cuneus, L Lingual Gyrus, L Superior Occipital Gyrus, R Superior Occipital Gyrus, R Middle Occipital Gyrus, R Inferior Occipital Gyrus, L Fusiform Gyrus, R Postcentral Gyrus, R Inferior Parietal Lobule, L Angular Gyrus, R Angular Gyrus, L Pallidum, R Pallidum, R Thalamus, R Heschl's Gyrus, L Superior Temporal Gyrus, R Temporal Pole, L Medial Temporal Pole, and R Medial Temporal Pole. These regions span prefrontal, cingulate, medial temporal, parietal/sensorimotor, thalamic/basal ganglia, and occipitotemporal visual systems; regional co-occurrence does not imply established inter-regional connectivity disruption.

\paragraph{Figshare MDD}
Under \(q_{\mathrm{BH}}<0.001\), the parameter family exhibiting the largest number of significant brain regions in Figshare MDD is \texttt{$\theta$} (23/90 AAL regions), followed by \texttt{$\tau_{i2}$} (18), \texttt{$C_4$} and \texttt{$r_{max}$} (13 each), and \texttt{$C_3$} and \texttt{$\beta$} (11 each). The 23 significant regions for \texttt{$\theta$} are:
L Superior Frontal Gyrus, L Superior Orbital Gyrus, L IFG (p. Triangularis), L IFG (p. Orbitalis), L Posterior-Medial Frontal, L Superior Medial Gyrus, L Cingulum Mid, R Hippocampus, R ParaHippocampal Gyrus, R Amygdala, R Superior Occipital Gyrus, L Fusiform Gyrus, R Fusiform Gyrus, R Postcentral Gyrus, L Paracentral Lobule, L Putamen, R Putamen, R Thalamus, L Heschl's Gyrus, R Heschl's Gyrus, L Middle Temporal Gyrus, L Inferior Temporal Gyrus, and R Inferior Temporal Gyrus. These regions span prefrontal, cingulate, limbic, temporal, occipital visual, sensorimotor, striatal, and thalamic systems.

\paragraph{Detailed Region-Wise Statistical Test Tables}
Given the modest sample size of PD31 where no parameter family surpassed the $q_{\mathrm{BH}}<0.001$ threshold with a large number of regions, we report the region-wise Welch's $t$-test statistics for the top-ranked parameter families in AD65 (\texttt{$C_1$}) and Figshare MDD (\texttt{$\theta$}) across all significant regions ($q_{\mathrm{BH}}<0.001$). Regions are ordered by AAL index ascendingly; $t$ denotes the Welch statistic (positive values indicate case mean exceeds control mean), $\mathrm{dof}$ is the Welch--Satterthwaite degrees of freedom, $p$ is the uncorrected two-tailed $p$-value, $q_{\mathrm{BH}}$ is the Benjamini--Hochberg FDR-adjusted $p$-value, and $|d|$ is the absolute Cohen's effect size.

\begin{longtable}{r p{0.36\textwidth} r r l l r}
\caption{Region-wise Welch's $t$-tests for the \texttt{C1} parameter family in the AD65 cohort (significant regions with $q_{\mathrm{BH}}<0.001$; $n_{\mathrm{AD}}=36$, $n_{\mathrm{control}}=29$).}
\label{tab:supp_ad65_c1_stats}\\
\toprule
ROI & AAL Brain Region & $t$ & dof & $p$ & $q_{\mathrm{BH}}$ & $|d|$ \\
\midrule
\endfirsthead
\multicolumn{7}{l}{\small Table~\ref{tab:supp_ad65_c1_stats} continued from previous page.}\\
\toprule
ROI & AAL Brain Region & $t$ & dof & $p$ & $q_{\mathrm{BH}}$ & $|d|$ \\
\midrule
\endhead
\bottomrule
\endfoot
2  & R Precentral Gyrus & 5.23 & 62.2 & $2.1\times10^{-6}$ & $4.3\times10^{-5}$ & 1.26 \\
3  & L Superior Frontal Gyrus & 4.69 & 63.0 & $1.5\times10^{-5}$ & $1.7\times10^{-4}$ & 1.14 \\
5  & L Superior Orbital Gyrus & 5.74 & 61.4 & $3.1\times10^{-7}$ & $1.2\times10^{-5}$ & 1.42 \\
6  & R Superior Orbital Gyrus & 5.11 & 62.6 & $3.2\times10^{-6}$ & $5.3\times10^{-5}$ & 1.24 \\
9  & L Middle Orbital Gyrus & 5.56 & 62.4 & $5.9\times10^{-7}$ & $2.0\times10^{-5}$ & 1.34 \\
11 & L IFG (p. Opercularis) & 4.36 & 46.8 & $7.0\times10^{-5}$ & $5.1\times10^{-4}$ & 1.14 \\
12 & R IFG (p. Opercularis) & 4.67 & 60.8 & $1.7\times10^{-5}$ & $1.9\times10^{-4}$ & 1.12 \\
13 & L IFG (p. Triangularis) & 5.23 & 62.3 & $2.1\times10^{-6}$ & $4.3\times10^{-5}$ & 1.29 \\
15 & L IFG (p. Orbitalis) & 5.88 & 62.9 & $1.7\times10^{-7}$ & $8.6\times10^{-6}$ & 1.43 \\
19 & L Posterior-Medial Frontal & 4.55 & 61.8 & $2.5\times10^{-5}$ & $2.4\times10^{-4}$ & 1.09 \\
21 & L Olfactory cortex & 5.33 & 62.8 & $1.4\times10^{-6}$ & $3.5\times10^{-5}$ & 1.29 \\
22 & R Olfactory cortex & 4.10 & 62.3 & $1.2\times10^{-4}$ & $7.8\times10^{-4}$ & 0.99 \\
29 & L Insula Lobe & 5.26 & 62.2 & $1.9\times10^{-6}$ & $4.1\times10^{-5}$ & 1.27 \\
31 & L Cingulum Ant & 5.68 & 63.0 & $3.7\times10^{-7}$ & $1.4\times10^{-5}$ & 1.38 \\
33 & L Cingulum Mid & 4.59 & 62.2 & $2.2\times10^{-5}$ & $2.2\times10^{-4}$ & 1.11 \\
37 & L Hippocampus & 5.17 & 62.5 & $2.6\times10^{-6}$ & $4.8\times10^{-5}$ & 1.25 \\
38 & R Hippocampus & 6.04 & 62.9 & $9.1\times10^{-8}$ & $6.4\times10^{-6}$ & 1.48 \\
39 & L ParaHippocampal Gyrus & 4.93 & 62.6 & $6.3\times10^{-6}$ & $9.2\times10^{-5}$ & 1.19 \\
45 & L Cuneus & 4.06 & 59.5 & $1.5\times10^{-4}$ & $9.0\times10^{-4}$ & 1.02 \\
47 & L Lingual Gyrus & 4.39 & 57.6 & $4.9\times10^{-5}$ & $4.0\times10^{-4}$ & 1.11 \\
49 & L Superior Occipital Gyrus & 4.02 & 62.9 & $1.6\times10^{-4}$ & $9.8\times10^{-4}$ & 0.98 \\
50 & R Superior Occipital Gyrus & 5.50 & 62.0 & $7.7\times10^{-7}$ & $2.3\times10^{-5}$ & 1.32 \\
52 & R Middle Occipital Gyrus & 5.02 & 62.0 & $4.7\times10^{-6}$ & $7.5\times10^{-5}$ & 1.21 \\
54 & R Inferior Occipital Gyrus & 4.91 & 62.2 & $7.1\times10^{-6}$ & $9.7\times10^{-5}$ & 1.18 \\
55 & L Fusiform Gyrus & 5.62 & 58.6 & $5.6\times10^{-7}$ & $1.9\times10^{-5}$ & 1.41 \\
58 & R Postcentral Gyrus & 5.27 & 63.0 & $1.8\times10^{-6}$ & $4.1\times10^{-5}$ & 1.29 \\
62 & R Inferior Parietal Lobule & 4.97 & 53.6 & $7.3\times10^{-6}$ & $9.9\times10^{-5}$ & 1.27 \\
65 & L Angular Gyrus & 4.43 & 62.3 & $3.9\times10^{-5}$ & $3.4\times10^{-4}$ & 1.09 \\
66 & R Angular Gyrus & 4.31 & 55.3 & $6.8\times10^{-5}$ & $5.0\times10^{-4}$ & 1.09 \\
75 & L Pallidum & 5.36 & 62.3 & $1.3\times10^{-6}$ & $3.4\times10^{-5}$ & 1.29 \\
76 & R Pallidum & 4.84 & 62.3 & $8.9\times10^{-6}$ & $1.2\times10^{-4}$ & 1.17 \\
78 & R Thalamus & 6.32 & 61.5 & $3.3\times10^{-8}$ & $4.1\times10^{-6}$ & 1.57 \\
80 & R Heschl's Gyrus & 4.38 & 63.0 & $4.7\times10^{-5}$ & $3.8\times10^{-4}$ & 1.06 \\
81 & L Superior Temporal Gyrus & 5.64 & 62.2 & $4.4\times10^{-7}$ & $1.6\times10^{-5}$ & 1.39 \\
84 & R Temporal Pole & 5.22 & 61.4 & $2.3\times10^{-6}$ & $4.3\times10^{-5}$ & 1.25 \\
87 & L Medial Temporal Pole & 4.14 & 60.3 & $1.1\times10^{-4}$ & $7.1\times10^{-4}$ & 1.03 \\
88 & R Medial Temporal Pole & 4.02 & 62.8 & $1.6\times10^{-4}$ & $9.8\times10^{-4}$ & 0.99 \\
\end{longtable}

\begin{longtable}{r p{0.36\textwidth} r r l l r}
\caption{Region-wise Welch's $t$-tests for the \texttt{$\theta$} parameter family in the Figshare MDD cohort (significant regions with $q_{\mathrm{BH}}<0.001$; $n_{\mathrm{MDD}}=35$, $n_{\mathrm{control}}=30$).}
\label{tab:supp_mdd_theta_stats}\\
\toprule
ROI & AAL Brain Region & $t$ & dof & $p$ & $q_{\mathrm{BH}}$ & $|d|$ \\
\midrule
\endfirsthead
\multicolumn{7}{l}{\small Table~\ref{tab:supp_mdd_theta_stats} continued from previous page.}\\
\toprule
ROI & AAL Brain Region & $t$ & dof & $p$ & $q_{\mathrm{BH}}$ & $|d|$ \\
\midrule
\endhead
\bottomrule
\endfoot
3  & L Superior Frontal Gyrus & -3.18 & 47.6 & $4.8\times10^{-6}$  & $1.0\times10^{-4}$ & 0.87 \\
5  & L Superior Orbital Gyrus & -6.27 & 44.7 & $1.3\times10^{-7}$ & $5.2\times10^{-6}$ & 1.76 \\
13 & L IFG (p. Triangularis) & -4.78 & 50.0 & $1.6\times10^{-5}$ & $2.6\times10^{-4}$ & 1.32 \\
15 & L IFG (p. Orbitalis) & -5.90 & 50.0 & $3.1\times10^{-7}$ & $9.7\times10^{-6}$ & 1.63 \\
19 & L Posterior-Medial Frontal & -5.96 & 45.6 & $3.4\times10^{-7}$ & $1.0\times10^{-5}$ & 1.67 \\
23 & L Superior Medial Gyrus & -5.78 & 42.5 & $7.8\times10^{-7}$ & $2.1\times10^{-5}$ & 1.63 \\
33 & L Cingulum Mid & -4.51 & 48.5 & $4.1\times10^{-5}$ & $4.8\times10^{-4}$ & 1.26 \\
38 & R Hippocampus & -6.49 & 49.3 & $4.0\times10^{-8}$ & $2.3\times10^{-6}$ & 1.81 \\
40 & R ParaHippocampal Gyrus & -4.41 & 49.4 & $5.5\times10^{-5}$ & $6.1\times10^{-4}$ & 1.22 \\
42 & R Amygdala & -4.78 & 40.8 & $2.3\times10^{-5}$ & $3.5\times10^{-4}$ & 1.35 \\
50 & R Superior Occipital Gyrus & -6.09 & 46.8 & $2.0\times10^{-7}$ & $6.8\times10^{-6}$ & 1.70 \\
55 & L Fusiform Gyrus & -7.33 & 46.9 & $2.7\times10^{-9}$ & $2.6\times10^{-7}$ & 2.05 \\
56 & R Fusiform Gyrus & -5.87 & 27.1 & $2.9\times10^{-6}$ & $7.1\times10^{-5}$ & 1.57 \\
58 & R Postcentral Gyrus & -4.43 & 48.1 & $5.4\times10^{-5}$ & $6.1\times10^{-4}$ & 1.24 \\
69 & L Paracentral Lobule & 4.99  & 41.1 & $1.1\times10^{-5}$ & $2.1\times10^{-4}$ & 1.41 \\
73 & L Putamen & -4.60 & 44.9 & $3.4\times10^{-5}$ & $4.3\times10^{-4}$ & 1.29 \\
74 & R Putamen & -5.17 & 40.5 & $6.6\times10^{-6}$ & $1.3\times10^{-4}$ & 1.46 \\
78 & R Thalamus & -6.71 & 44.1 & $3.0\times10^{-8}$ & $1.8\times10^{-6}$ & 1.88 \\
79 & L Heschl's Gyrus & -4.82 & 49.0 & $1.4\times10^{-5}$ & $2.3\times10^{-4}$ & 1.34 \\
80 & R Heschl's Gyrus & -6.88 & 39.6 & $2.9\times10^{-8}$ & $1.8\times10^{-6}$ & 1.94 \\
85 & L Middle Temporal Gyrus & -4.33 & 41.2 & $9.4\times10^{-5}$ & $8.7\times10^{-4}$ & 1.18 \\
89 & L Inferior Temporal Gyrus & -5.19 & 42.3 & $5.7\times10^{-6}$ & $1.2\times10^{-4}$ & 1.46 \\
90 & R Inferior Temporal Gyrus & 6.61  & 43.9 & $4.3\times10^{-8}$ & $2.4\times10^{-6}$ & 1.80 \\
\end{longtable}


\bibliographystyle{unsrtnat}
\bibliography{references}